\def\year{2022}\relax
\documentclass[letterpaper]{article} % DO NOT CHANGE THIS
\usepackage{aaai22}  % DO NOT CHANGE THIS
\usepackage{times}  % DO NOT CHANGE THIS
\usepackage{helvet}  % DO NOT CHANGE THIS
\usepackage{courier}  % DO NOT CHANGE THIS
\usepackage[hyphens]{url}  % DO NOT CHANGE THIS
\usepackage{graphicx} % DO NOT CHANGE THIS

\usepackage{natbib}  % DO NOT CHANGE THIS AND DO NOT ADD ANY OPTIONS TO IT
\usepackage{caption} % DO NOT CHANGE THIS AND DO NOT ADD ANY OPTIONS TO IT
\DeclareCaptionStyle{ruled}{labelfont=normalfont,labelsep=colon,strut=off} % DO NOT CHANGE THIS
\usepackage{algorithm}
\usepackage{algorithmic}

\usepackage{amsfonts,amssymb}
\usepackage{amsmath} %\boldsymbol
\usepackage{colortbl}
\usepackage[dvipsnames]{xcolor}
\usepackage{subfigure}
\usepackage{multirow}
\usepackage{newfloat}
\usepackage{listings}
\floatstyle{ruled}
\newfloat{listing}{tb}{lst}{}
\floatname{listing}{Listing}
\title{Molecular Contrastive Learning with Chemical Element Knowledge Graph}
\author{
    Yin Fang\textsuperscript{\rm 1,\rm2,\rm 3}\equalcontrib, Qiang Zhang\textsuperscript{\rm 1,\rm2,\rm 3}\equalcontrib, Haihong Yang\textsuperscript{\rm 1,\rm2,\rm 3}, Xiang Zhuang\textsuperscript{\rm 1,\rm2,\rm 3}, Shumin Deng\textsuperscript{\rm1,\rm3},
    Wen Zhang\textsuperscript{\rm 4}, Ming Qin\textsuperscript{\rm 1,\rm2,\rm 3}, Zhuo Chen\textsuperscript{\rm 1,\rm2,\rm 3}, Xiaohui Fan\textsuperscript{\rm 5,\rm6,\rm 7}, Huajun Chen\textsuperscript{\rm 1,\rm2,\rm 3}\thanks{Corresponding Author.}
}
\affiliations{
    \textsuperscript{\rm 1} College of Computer Science and Technology, Zhejiang University \\
    \textsuperscript{\rm 2} ZJU-Hangzhou Global Scientific and Technological Innovation Center, Zhejiang University \\
    \textsuperscript{\rm 3} AZFT Joint Lab for Knowledge Engine 
     \textsuperscript{\rm 4} School of Software Technology, Zhejiang University \\
    \textsuperscript{\rm 5} Pharmaceutical Informatics Institute, College of Pharmaceutical Sciences, Zhejiang University \\
    \textsuperscript{\rm 6} Innovation Center in Zhejiang University, State Key Laboratory of Component-Based Chinese Medicine\\
    \textsuperscript{\rm 7} Westlake Laboratory of Life Sciences and Biomedicine

    \{fangyin, qiang.zhang.cs, haihong825, zhuangxiang, 231sm, wenzhang2015, \\
    qinandming, zhuo.chen, fanxh, huajunsir\}@zju.edu.cn
}

\usepackage{bibentry}
\begin{document}

\maketitle

\begin{abstract}
Molecular representation learning contributes to multiple downstream tasks such as molecular property prediction and drug design. To properly represent molecules, graph contrastive learning is a promising paradigm as it utilizes self-supervision signals and has no requirements for human annotations. However, prior works fail to incorporate fundamental domain knowledge into graph semantics and thus ignore the correlations between atoms that have common attributes but are not directly connected by bonds. To address these issues, we construct a Chemical Element Knowledge Graph (KG) to summarize microscopic associations between elements and propose a novel \textbf{K}nowledge-enhanced \textbf{C}ontrastive \textbf{L}earning (\textbf{KCL}) framework for molecular representation learning.
KCL framework consists of three modules. 
The first module, \textit{knowledge-guided graph augmentation}, augments the original molecular graph based on the Chemical Element KG. 
The second module, \textit{knowledge-aware graph representation}, extracts molecular representations with a common graph encoder for the original molecular graph and a  Knowledge-aware Message Passing Neural Network (KMPNN) to encode complex information in the augmented molecular graph. 
The final module is a \textit{contrastive objective}, where we maximize agreement between these two views of molecular graphs.
% and disagreement between hard negatives.
%By maximizing the agreement of the knowledge-enhanced augmented graph and the original molecular graph, KCL can incorporate both domain knowledge and structural topology into molecular representations. 
Extensive experiments demonstrated that KCL obtained superior performances against state-of-the-art baselines on eight molecular datasets. Visualization experiments properly interpret what KCL has learned from atoms and attributes in the augmented molecular graphs. Our codes and data are available at https://github.com/ZJU-Fangyin/KCL. 
\end{abstract}

\section{Introduction}
Accurately predicting the properties of molecules lies at the core of fundamental tasks in the chemical and pharmaceutical communities. In light of deep learning, several supervised models have been investigated to learn molecular representations through predicting molecular properties~\cite{DBLP:conf/icml/GilmerSRVD17,DBLP:journals/jcisd/YangSJCEGGHKMPS19,DBLP:conf/ijcai/SongZNFLY20}. While effective, these methods face the challenges of limited labeled data, as laboratory experiments are expensive and time-consuming to annotate data. Moreover, due to the enormous diversity of chemical molecules, these works could barely generalize to unseen cases~\cite{DBLP:conf/iclr/HuLGZLPL20,DBLP:conf/nips/RongBXX0HH20}, which greatly hinders practical applicability.

One line of works to alleviate these issues is to design pretext tasks to learn node or graph representations without labels. Several attempts have been made to investigate different strategies for such tasks, including masked attribute prediction~\cite{DBLP:conf/iclr/HuLGZLPL20}, graph-level motif prediction~\cite{DBLP:conf/nips/RongBXX0HH20}, and graph context prediction~\cite{DBLP:conf/nips/LiuDL19}. The other line follows a contrastive learning framework from the computer vision domain~\cite{DBLP:conf/cvpr/WuXYL18,DBLP:conf/icml/ChenK0H20}, which aims to construct similar and dissimilar view pairs via graph augmentations, including node dropping, edge perturbation, subgraph extraction, and attribute masking~\cite{DBLP:conf/nips/YouCSCWS20}. Due to the smaller amount of parameters and simpler predefined tasks, we adopt contrastive learning in our work.

However, unlike images, contrastive learning on graphs has its unique challenges. First, the structural information and semantics of the graphs vary significantly across domains, which makes it difficult to design a universal augmentation scheme for graphs. Especially for molecular graphs, removing or adding a chemical bond or a functional group will drastically change their identities and properties~\cite{DBLP:conf/nips/YouCSCWS20}. More importantly, existing graph contrastive learning models mainly focus on graphs structures, without considering fundamental domain knowledge into graph semantics. Another neglected defect is that they model the atoms in molecular graphs as individuals that can only interact when there exists an edge (i.e., a chemical bond), 
failing to consider the correlations between atoms (e.g., commonalities between atoms of the same attributes).

\begin{figure}%[!b]
\centering
\includegraphics[width=1\columnwidth]{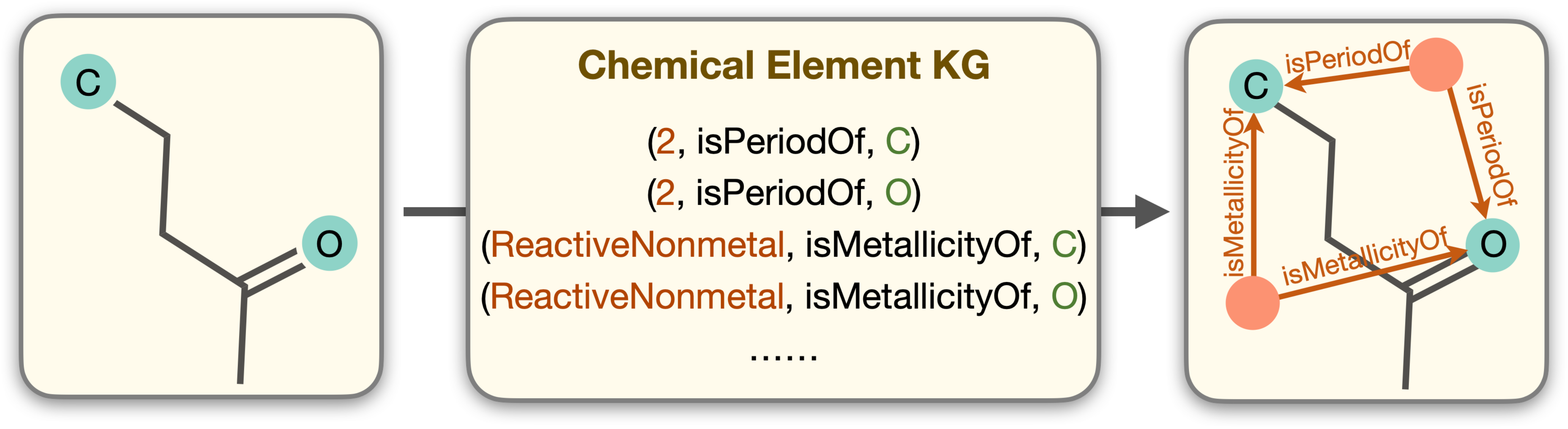} % Reduce the figure size so that it is slightly narrower than the column. Don't use precise values for figure width.This setup will avoid overfull boxes.
\caption{Chemical Element KG builds associations between atoms that are not directly connected by bonds but related in fundamental chemical attributes, as denoted by red arrows.}
\label{intro}
\end{figure}

To overcome these challenges, we enrich the molecular graph contrastive learning by incorporating domain knowledge. Since chemical domain knowledge is crucial prior, we hypothesize that the attributes of elements (atom is an instance of element) can affect molecular properties. To obtain the domain knowledge and build microscopic correlations between atoms, we first construct a Chemical Element Knowledge Graph (KG) based on Periodic Table of Elements~\footnote{\url{https://ptable.com}}.  
The Chemical Element KG describes the relations between elements (denoted in green in Figure~\ref{intro}) and their basic chemical attributes (e.g., periodicity and metallicity, denoted in red in Figure~\ref{intro}). Then we augment the original molecular graph with the guidance of Chemical Element KG, as shown in Figure~\ref{intro}, which helps to establish the associations between atoms that have common attributes but are not directly connected by bonds. In this way, the augmented molecular graph contains not only structural topologies but also the fundamental domain knowledge of elements.

On top of that, we propose a novel \textbf{K}nowledge-enhanced \textbf{C}ontrastive \textbf{L}earning (KCL) framework to improve the molecular representation with three modules. {(1)} The \textit{knowledge-guided graph augmentation} module leverages Chemical Element KG to guide the graph augmentation process. While preserving the topology structure, the augmented molecular graph also builds associations that cannot be observed explicitly. {(2)} The \textit{knowledge-aware graph representation} module learns molecular representations. We adopt a commonly used graph encoder for the original molecular graphs while designing a Knowledge-aware Message Passing Neural Network (KMPNN) encoder to provide heterogeneous attentive message passing for different types of knowledge in the augmented molecular graph. {(3)} The \textit{contrastive objective} module trains the encoders to maximize the agreement between positives and the discrepancy between hard negatives. To the best of our knowledge, it is the first work to construct KG based on fundamental knowledge of chemical elements and guide molecular contrastive learning. 
%Chemical Element KG can also be extended to other categories of tasks that require fundamental element knowledge in chemical and biological domains.
%
Our contributions can be summarized as follows:
\begin{itemize}
\item We construct a Chemical Element KG, which describes the relations between elements and their chemical attributes. It can assist various molecular learning tasks beyond the ones in this paper.
\item We develop a new contrastive learning framework (KCL) with three modules: knowledge-guided graph augmentation, knowledge-aware graph representation, and contrastive objective.
% We propose a new contrastive learning framework (KCL) with three technical highlights. (i) We propose a knowledge-guided augmentation scheme for molecular graphs based on the Chemical Element KG such that the microscopic correlations between atoms can be considered. (ii) We introduce a new graph-based encoder, KMPNN, which provides heterogeneous attentive message passing for different type of neighbors. (iii) We design a hard negative sampling strategy that treats molecules similar to the anchor instance in the fingerprint space as negatives.
\item  We evaluate KCL on eight various molecular datasets under both fine-tune and linear protocols and demonstrate its superiority over the state-of-the-art methods.
\end{itemize}

\section{Related Works}
\paragraph{Molecular Representation Learning}
%Molecular representation learning has been one of the core topics in chemical property prediction and drug discovery, whose goal is to embed a molecule into a low-dimensional vector space. Descriptor-based methods (such as traditional chemical fingerprints) encode the neighbors of each atom into a fixed bit string with a hash function. 
In light of deep learning,~\citeauthor{DBLP:conf/nips/DuvenaudMABHAA15} first applied convolutional networks to map molecules into neural fingerprints. Subsequent works fed SMILES (a line notation for describing the structure of chemical species using short ASCII strings) into recurrent networks-based models to produce molecular representations~\cite{DBLP:journals/corr/JastrzebskiLC16,DBLP:conf/bcb/0005WZH17}. To utilize topology information in the molecular graph, MPNN~\cite{DBLP:conf/icml/GilmerSRVD17} and its variants DMPNN~\cite{DBLP:journals/jcisd/YangSJCEGGHKMPS19}, CMPNN~\cite{DBLP:conf/ijcai/SongZNFLY20}, CoMPT~\cite{DBLP:journals/corr/abs-2107-08773} leverage the node and edge attributes during message passing. 
% DMPNN~\cite{DBLP:journals/jcisd/YangSJCEGGHKMPS19} treated the molecular graph as an edge-oriented directed structure, avoiding the information redundancy during iterations. Following the edge-based message passing framework,
% CMPNN~\cite{DBLP:conf/ijcai/SongZNFLY20} introduced the node-edge interaction module to leverage the node and edge attributes during message passing. CoMPT~\cite{DBLP:journals/corr/abs-2107-08773} invokes a communicative message-passing paradigm based on Transformer. 
However, all the above-mentioned works are supervised models, require expensive annotations, and could barely generalize to unseen molecules, which greatly hinders the feasibility in practice.

\paragraph{Self-Supervised Learning on Graphs}
Self-supervised learning addresses such a limitation by pre-training molecular graphs.
~\citeauthor{DBLP:conf/nips/LiuDL19} exploited the idea of N-gram in NLP and conducted vertices embedding by predicting the vertices attributes.
~\citeauthor{DBLP:conf/iclr/HuLGZLPL20} designed two pre-training tasks, i.e., predicting neighborhood context and node attributes, to learn meaningful node representations, then using graph-level multi-task pre-training to refine graph representations. Alternatively, GROVER~\cite{DBLP:conf/nips/RongBXX0HH20} incorporated a Transformer-style architecture and learned node embeddings by predicting contextual properties and motif information. Other works~\cite{DBLP:conf/ijcai/ShangMXS19,DBLP:conf/aaai/SunLZ20,DBLP:conf/icml/YasunagaL20} utilized similar strategies for either node or graph level pre-training.% in the context of a more specific task or domain.

\paragraph{Contrastive Learning on Graphs}
Contrastive learning is a widely-used self-supervised learning algorithm. Its main idea is to make representations of positive pairs that agree with each other and negatives disagree as much as possible ~\cite{DBLP:conf/nips/YouCSCWS20}. One key component is to generate informative and diverse views from each data instance. 
% Early efforts utilize nodes as views for graph contrastive learning, but they are less efficient at capturing whole-graph characteristics. 
Previous graph augmentations generated views by randomly shuffling node features~\cite{DBLP:conf/iclr/VelickovicFHLBH19,DBLP:conf/icml/HassaniA20}, removing edges or masking nodes~\cite{DBLP:conf/nips/YouCSCWS20}. However, these perturbations may hurt the domain knowledge inside graphs, especially for chemical compounds. MoCL~\cite{DBLP:journals/corr/abs-2106-04509} proposed a substructure substitution and incorporated two-level knowledge to learn richer representations, CKGNN~\cite{DBLP:journals/corr/abs-2103-13047} selected positive pairs via fingerprint similarity. But they ignore the fundamental domain knowledge. 
% Moreover, recent works from the vision and language domains~\cite{DBLP:conf/cvpr/SchroffKP15,DBLP:conf/iclr/RobinsonCSJ21,DBLP:journals/corr/abs-2104-08821} argue that contrastive learning benefits from hard negative samples. Hence we design a hard negative mining strategy for molecular graphs.
%~\cite{zhu2021empirical} measure hardness of each negative pair and upweights hard negative samples. \add{While representation-similar-graphs during training process may not be true hard negatives.}

\section{Methodology}
\begin{figure*}[!ht]
\centering
\includegraphics[width=2\columnwidth]{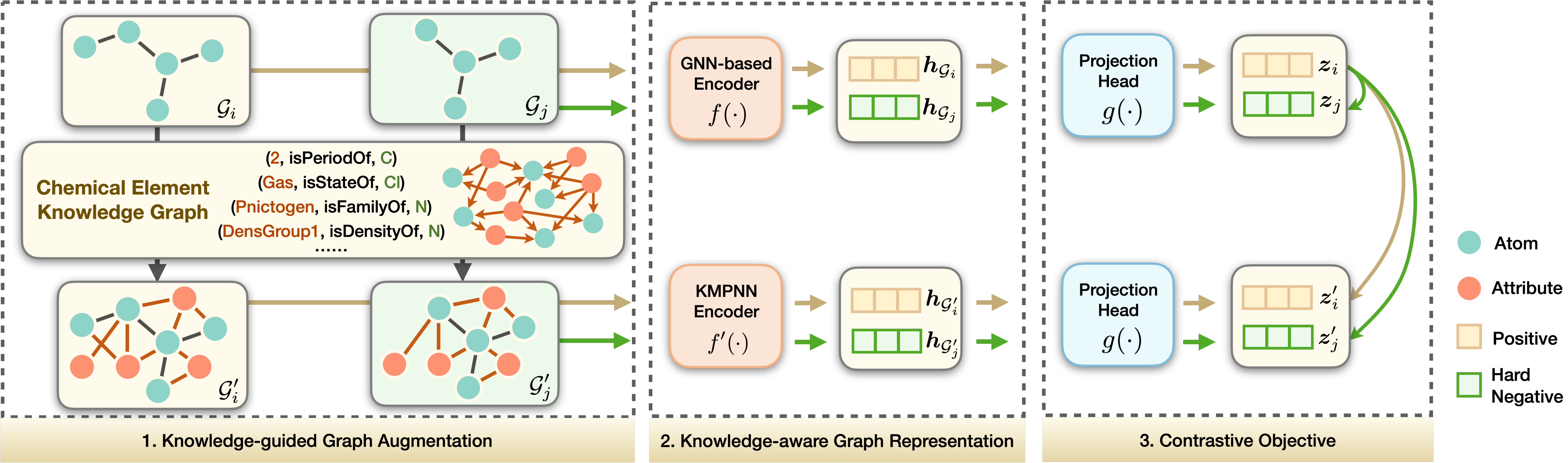} % Reduce the figure size so that it is slightly narrower than the column. Don't use precise values for figure width.This setup will avoid overfull boxes.
\caption{An illustrative example for KCL. We ignore edge directions in four molecular graphs due to space limitation (the direction of an edge between an attribute and an atom is from the former to the latter, while an edge between atoms is bidirectional). Module 1: Knowledge-guided graph augmentation converts the original molecular graph $\mathcal{G}$ into the augmented molecular graph $\mathcal{G}'$ based on Chemical Element KG. Module 2: Knowledge-aware graph representation captures representations from two graph views separately. Module 3: Contrastive objective trains the encoders and the projection head to maximize agreement between positives and disagreement between hard negatives (e.g., $\mathcal{G}_j$ act as the hard negative of $\mathcal{G}_i$) via a contrastive loss.}
\label{overview}
\vspace{-1em}
\end{figure*}

\subsection{Problem Formulation}
A molecule can be represented as a graph $\mathcal{G}=\{\mathcal{V}, \mathcal{E}\}$, where $|\mathcal{V}|$ denotes a set of $n$ atoms (nodes) and $|\mathcal{E}| $ denotes a set of $m$ bonds (edges). Each edge is bidirectional. 
Let $\mathcal{N}_v$ denote the set of node $v$'s neighbors. We use $x_v$ to represent the initial features of node $v$, and $e_{uv}$ as the initial features of edge $(u,v)$. Let $\boldsymbol{h}(v)$ be the node hidden state and $\boldsymbol{h}(e_{uv})$ for the edge hidden state. 
%We randomly initialize embeddings of each type of atoms and bonds.
% We further denote the node feature matrix, the adjacency matrix and edge feature matrix as $\boldsymbol{X} \in \mathbb{R}^{|\mathcal{V}| \times d_{1}}$, $\boldsymbol{A} \in \mathbb{R}^{|\mathcal{V}| \times |\mathcal{V}|}$ and $\boldsymbol{E} \in \mathbb{R}^{|\mathcal{E}| \times d_{2}}$. 
In the setting of self-supervised graph representation learning, our goal is to learn graph encoders $f: \mathcal{G} \mapsto \mathbb{R}^d$ which maps an input graph to a vector representation without the presence of any labels. The learned encoders and representations can then be used for downstream tasks.
% via the fine-tune or linear classifier.
% with evaluation details in Experiments. 
% \add{$f: \mathcal{G} \mapsto \mathbb{R}^d$}

% The molecular property prediction task can be defined as a classification/regression problem on a molecular graph ${G}$. 
%, the task is to predict the label ${y}$. 

\subsection{Overview}
Figure~\ref{overview} shows the overview of our work. We propose a contrastive learning framework called KCL with three modules: (1) Knowledge-guided graph augmentation transforms any given molecule graph $\mathcal{G}$ into an augmented molecular graph $\mathcal{G}'$ with the guidance of Chemical Element KG. (2) Knowledge-aware graph representation aims to extract representations from $\mathcal{G}$ and $\mathcal{G}'$ respectively. (3) Contrastive objective aims to project representations to the space where contrastive loss is applied and train the encoders to maximize the agreement between positive pairs and the discrepancy between hard negatives. 

% \subsection{Overview}
% In this paper, we propose a contrastive learning framework called KCL with three modules: (1) \textit{Knowledge-guided Graph Augmentation}, (2) \textit{Pluggable Graph Encoders}, and (3) \textit{Knowledge-enhanced Contrastive Objective}. Figure~\ref{overview} shows the key idea of the three modules.
% %
% The knowledge-guided graph augmentation module transforms any given molecule graph $G$ into a knowledge-enhanced augmented graph $G'$ with the assistance of Chemical Element KG.
% %
% Pluggable graph encoders $f(\cdot)$ and $f'(\cdot)$ aim to extract representations from $G$ and $G'$ respectively. Our framework allows various choices of the network architecture for graph encoders. Here we utilize a GNN-based encoder for $G$ and design a KMPNN encoder to encode complex information in the augmented graph $G'$.
% %
% Finally, the knowledge-enhanced contrastive objective aims to projects representations to the space where a contrastive loss is applied and the encoders are trained to maximize the agreement between positive pairs and the discrepancy between hard negatives. 

\subsection{Knowledge-guided Graph Augmentation}
\paragraph{Chemical Element KG Construction.}
The prerequisite of our work is to collect the fundamental chemical domain knowledge. Previous attempts~\cite{delmas2021building,DBLP:conf/ijcai/LinQWMZ20} built KGs from the public chemical database and scientific literature to extract associations between chemicals and diseases or drug pairs, but none of them considered the fundamental information in chemical elements. In contrast, we crawl all the chemical elements and their attributes from the Periodic Table of Elements. Each element contains more than 15 attributes, including metallicity, periodicity, state, weight, electronegativity, electron affinity, melting point, boiling point, ionization, radius, hardness, modulus, density, conductivity, heat, and abundance.

After that, the extracted triples in the form of (Gas, isStateOf, Cl) are constructed in KG, indicating that there are specified relations between elements and attributes. However, since each element has some different continuous attributes, it is difficult for KG to model their connections. To overcome this difficulty, we histogramize the continuous attributes and convert them into discrete labels (e.g., DensityGroup1, RadiusGroup2). The statistics of Chemical Element KG are summarized in Table~\ref{KGstat}.
% Then we convert the raw data into triples in the form of $(h,r,t)$ where $h$, $r$ and $t$ refers to the head entity, relation, and tail entity respectively. 
% In general, KG is inherently described as a heterogeneous network, and can provide abundant information with structural relations among entities, and unstructured semantic relations associated with each entity. 
%Since the periodic table of elements contains the fundamental chemical knowledge, we convert it to Element KG to store micro-level domain knowledge of elements. 

\begin{table}[!t]
\small
\centering
%\resizebox{.95\columnwidth}{!}{
\begin{tabular}{lcl}
\hline
\hline
     & \small{Chemical Element KG} \\
\hline
Elements & 118\\
Attributes & 107\\
Entities & 225\\
Relation Types & 17\\
KG Triples & 1643\\
\hline
\hline
\end{tabular}
\caption{The statistics of Chemical Element KG.}
\label{KGstat}
\vspace{-1em}
\end{table}

%To incorporate fundamental domain knowledge, we consider neighborhood topologies for atom related entities, in the form of knowledge graph. 
\paragraph{Graph Augmentation.}
Since most existing augmentation approaches (e.g., node dropping and edge perturbation) violate the chemical semantic inside molecules and ignore the influence of fundamental knowledge on graph semantics, we address these issues by proposing a knowledge-guided graph augmentation module with the guidance of Chemical Element KG. Specifically, as shown in Figure~\ref{overview}, we extract 1-hop neighbor attributes (nodes in red) of atoms (nodes in green) in a molecule from Chemical Element KG and add the triples as edges (edges in red). For example, we add a node ``Gas'' and an edge from ``Gas'' to ``Cl'' to the original molecular graph based on the triple (Gas, isStateOf, Cl).
Note that the direction of each edge between the attribute and the atom is from the former to the latter, while the edges between atoms are bidirectional. Then we obtain an augmented molecular graph, in which the original molecular structure is preserved, and neighborhood topologies for atom-related attributes are introduced.

% Formally, we denote the augmented graph by $\widetilde G=\{\mathcal{\widetilde V}, \mathcal{\widetilde E}\}$, where $\mathcal{\widetilde V}$ is the set of atoms and attributes, $\mathcal{\widetilde E}$ is the set of chemical bonds and atom relations.

%Based on the knowledge-guided graph augmentation, 
%the molecular graph $\mathcal{G}$ is converted to the 
While preserving the topology structure, the augmented molecular graph $\mathcal{G}'$ also considers the fundamental domain knowledge within elements, as well as the microscopic associations between atoms that have common attributes but are not directly connected by bonds. The augmented molecular graph thus contains richer and more complex information, and is treated as a positive sample in contrastive learning.

\subsection{Knowledge-aware Graph Representation}
\paragraph{Knowledge Feature Initialization.}
Different from the random initialization of atoms and bonds, in order to obtain the initial features of attributes and relations in the augmented molecular graph, we adopt the commonly used KG embedding method, RotateE~\cite{DBLP:conf/iclr/SunDNT19}, to train Chemical Element KG. In this way, the initial features can capture the structural information of the triples. The necessity of this step is proved in subsequent experiments. 
More details are in ~\ref{Feature Initialization} of Appendix. 
% In addition, we randomly initialize the embeddings of each type of atoms and bonds.

% Our framework allows various choices of network architecture for graph encoders. Here we utilize a GNN-based encoder for $\mathcal{G}$ and design a KMPNN encoder for $\mathcal{G}'$.

\paragraph{KMPNN Encoder.} Although various architectures can be adopted, since the augmented molecular graphs are complex irregular-structured data that combines two types of information (i.e., the structure knowledge implied in molecular bonds and domain knowledge extracted from Chemical Element KG), we design a KMPNN encoder as $f'(\cdot)$ to learn their graph-level representations. The key idea behind KMPNN is that we provide two types of message passing for different types of neighbors, and assign them different attention according to their importance. 
%map nodes to real-value embedding vectors in a low-dimensional space and then convert them to a graph-level representation with a readout function.

Algorithm~\ref{algorithm:KMPNN} describes the KMPNN encoding process. The input of the encoder is the augmented molecular graph $\mathcal{G}'=\{\mathcal{V}, \mathcal{E}\}$, including initial features of all nodes $x_v$, $\forall v \in \mathcal{V}$, and features of all edges $e_{uv}$, $\forall (u,v) \in \mathcal{E}$. $K$ rounds of message passing are then applied to all nodes. We enable heterogeneous message passing with two $\mathrm{M}\scriptstyle\mathrm{SG}$ functions, where $\mathrm{M}\scriptstyle\rm{SG_1\displaystyle(\cdot)}$ is applied to neighbors representing atoms, and $\mathrm{M}\scriptstyle\rm{SG_0\displaystyle(\cdot)}$ is applied to attributes in the neighborhood. The indicator function $\boldsymbol{1}_{[\cdot]}$ is used to index the selection of these functions, $\boldsymbol{1}_{[u=a]}=1$ if $u$ represents an atom else 0. In this way, the nodes with the same type of knowledge share parameters during message passing. 

Apart from the above, we extend message passing by self-attention. We compute attention coefficients and normalize them with the softmax function to make coefficients easily comparable across different nodes.
% \begin{small}
% \begin{equation}
% \alpha_{uv}=\mathrm{softmax} \left(\left(a\left(\boldsymbol{W} \boldsymbol{h}_{u}, \boldsymbol{W} \boldsymbol{h}_{v}\right)\right)\right).
% \end{equation}
% \end{small}
% Here, $a$ is a shared attentional mechanism and $\boldsymbol{W}$ is a weight matrix which acts as a shared linear transformation, \small$a\left(\boldsymbol{W} \boldsymbol{h}_{u}, \boldsymbol{W} \boldsymbol{h}_{v}\right)$ \normalsize indicates the importance of node $u$'s features to node $v$. 
Following ~\cite{DBLP:conf/iclr/VelickovicCCRLB18}, the coefficients can be expressed as:
\begin{small}
\begin{equation}
\alpha_{uv}=\frac{\exp \left(\mathrm{LeakyReLU}\left(\boldsymbol{a}^T\left[\boldsymbol{W} \boldsymbol{h}_{u}|| \boldsymbol{W} \boldsymbol{h}_{v}\right]\right)\right)}{\sum_{k \in \mathcal{N}_{u}} \exp \left(\mathrm{LeakyReLU}\left(\boldsymbol{a}^T\left[\boldsymbol{W} \boldsymbol{h}_{u}|| \boldsymbol{W} \boldsymbol{h}_{k}\right]\right)\right)},
\end{equation}
\end{small}
where $\cdot ^T$ represents transposition and $||$ is the concatenation operation. The attention mechanism is implemented as a single-layer feedforward neural network, parametrized by a weight vector $\boldsymbol{a}$ and followed by a LeakyRELU activation.

Once obtained, the normalized attention coefficients are used to compute a linear combination of the features corresponding to them:
\begin{small}
\begin{equation}
\mathrm{M}\scriptstyle\mathrm{SG}_{0}\displaystyle= \alpha_{uv}\boldsymbol{W}_0\boldsymbol{h}^{k-1}(e_{uv}) \cdot \boldsymbol{h}^{k-1}(u),
\end{equation}
\end{small}
where $\boldsymbol{W}_0$ denotes the weight matrix operating on incoming relations. This attentive message passing function allows for assigning different attention values to neighbor nodes, based on the intuition that different attributes have different importance to the atom.

\begin{algorithm}[!t]
\caption{KMPNN encoding algorithm.}
\label{algorithm:KMPNN}
\textbf{Input}: The augmented molecular graph $\mathcal{G}'=\{\mathcal{V}, \mathcal{E}\}$; message function $\mathrm{M}\scriptstyle\mathrm{SG\displaystyle(\cdot)}$; 
%indicator function $\boldsymbol{1}[u=a]=1$ if $u$ is an atom else 0;
%depth $K$; input node and edge features \{$x_v, \forall v \in \mathcal{V_E},e_{uv},\forall (u,v) \in \mathcal{R}$\}; 
%trainable functions $\mathrm{M}\scriptstyle\mathrm{SG_1\displaystyle(\cdot)}$ for atom neighbors,  $\mathrm{M}\scriptstyle\rm{SG_0\displaystyle(\cdot)}$ for the rest of nodes in the neighborhood; 
aggregate function $\mathrm{A}\scriptstyle\mathrm{GG}$; update function $\mathrm{U}$.  \\
%readout function $\mathrm{R}\scriptstyle\mathrm{EADOUT}$.\\
%\textbf{Parameter}: Optional list of parameters\\
\textbf{Output}: Graph embedding $\boldsymbol{h}_{\mathcal{G}'}$.
\begin{algorithmic}[1] %[1] enables line numbers
\STATE  $\boldsymbol{h}^0(v)\leftarrow x_v$, $\forall v \in \mathcal{V}$; $\boldsymbol{h}^0(e_{uv})\leftarrow e_{uv}$, $\forall (u,v) \in \mathcal{E}$
\FOR{$k=1,\dots,K$}
\FOR{$v \in \mathcal{V}$}
\STATE 
$\boldsymbol{m}^k(v) \leftarrow  \mathrm{A}\scriptstyle\mathrm{GG}\displaystyle(\{\mathrm{M}\scriptstyle\mathrm{SG}_{\boldsymbol{1}[u=a]}\displaystyle(\boldsymbol{h}^{k-1}(e_{uv}),\boldsymbol{h}^{k-1}(u)), $ $\forall u \in \mathcal{N}(v)\})$ \\
$\boldsymbol{h}^k(v) \leftarrow \mathrm{U}\displaystyle(\boldsymbol{h}^{k-1}(v), \boldsymbol{m}^k(v))$
\ENDFOR
\ENDFOR
\STATE $\boldsymbol{h}_{\mathcal{G}'} \leftarrow \mathrm{R}\scriptstyle\mathrm{EADOUT}\displaystyle (\{\boldsymbol{h}^{K}(v) ,\forall v \in \mathcal{V} \})$
\end{algorithmic}
\end{algorithm}

Since the messages delivered by different neighbor atoms to the central atom also have various importance, atoms in the neighborhood follow a common process with different parameters:
% \begin{small}
% \begin{equation}
% \beta_{uv}=\frac{\exp \left(\mathrm{LeakyReLU}\left(\boldsymbol{a'}^T\left[\boldsymbol{W'} \boldsymbol{h}_{u}|| \boldsymbol{W'} \boldsymbol{h}_{v}\right]\right)\right)}{\sum_{k \in \mathcal{N}_{u}} \exp \left(\mathrm{LeakyReLU}\left(\boldsymbol{a'}^T\left[\boldsymbol{W'} \boldsymbol{h}_{u}|| \boldsymbol{W'} \boldsymbol{h}_{k}\right]\right)\right)}, \\
% \end{equation}
% \end{small}
\begin{small}
\begin{equation}
\mathrm{M}\scriptstyle\mathrm{SG}_{1}\displaystyle
=\beta_{uv}\boldsymbol{W}_1\boldsymbol{h}^{k-1}(e_{uv}) \cdot \boldsymbol{h}^{k-1}(u),
\end{equation}
\end{small}
where $\beta_{uv}$ is the attention coefficients between atoms, $\boldsymbol{W_1}$ is the weight matrix of incoming bonds. 

% \paragraph{Message Diffusion and Global Pooling.}
In the message diffusion module, we collect the messages from their neighboring edges in message aggregation, 
\begin{small}
\begin{equation}
\boldsymbol{m}^k(v)= \sum_{k \in \mathcal{N}_{u}} \mathrm{M}\scriptstyle\mathrm{SG}\displaystyle(\boldsymbol{h}^{k-1}(e_{uv}) , \boldsymbol{h}^{k-1}(u)),
\end{equation}
\end{small}
and apply GRU as the update function.
\begin{small}
\begin{equation}
\boldsymbol{h}^{k}(v) =\mathrm{GRU}(\boldsymbol{h}^{k-1}(v), \boldsymbol{m}^k(v)),
\end{equation}
\end{small}
where GRU is the Gated Recurrent Unit introduced in~\cite{DBLP:conf/ssst/ChoMBB14}. After $K$ steps' iteration, a readout operator is applied to get a graph-level representation for the molecule:
\begin{small} 
\begin{equation}
\boldsymbol{h}_{\mathcal{G}'} = \mathrm{Set2set}(\boldsymbol{h}^{K}(v)),
\end{equation}
\end{small}
where set2set~\cite{DBLP:journals/corr/VinyalsBK15} is specifically designed to operate on sets and have more expressive power than simply summing the final node states.

%Subsequent experimental verification shows that we make better use of knowledge from different sources with this knowledge-aware message passing scheme.
%\subsection{Knowledge-Enhanced Contrastive Learning}

%We propose KCL for (self-supervised) graph pre-training. Similar to conventional contrastive learning, KCL pre-trains graphs by maximizing the agreement between two views of the same graph and minimize the agreement of two views of different graphs via a contrastive loss in the latent space. The framework consists of the following five major components as shown in Figure~\ref{overview}.
\paragraph{GNN-based Encoder.} There is no constraint of network architecture for $f(\cdot)$. We opt for simplicity and adopt the commonly used GCN~\cite{DBLP:conf/iclr/KipfW17} to obtain $\boldsymbol{h}_\mathcal{G}=f(\mathcal{G})$, which is the output after weighted sum and max pooling readout. 

\subsection{Contrastive Objective}
\paragraph{Projection Head.}
A non-linear transformation $g(\cdot)$ named projection head maps both the original and augmented representations to another latent space where the contrastive loss is calculated, as advocated in~\cite{DBLP:conf/icml/ChenK0H20}. In KCL, a two-layer perceptron (MLP) is applied to obtain $\boldsymbol{z}=g(\boldsymbol{h}_{\mathcal{G}})$ and $\boldsymbol{z'}=g(\boldsymbol{h}_{\mathcal{G}'})$. Note that after pre-training is completed, we throw the projection head away and only use the encoders for downstream tasks.

\paragraph{Negative Mining.}
% Different from previous graph contrastive learning works that randomly choose graphs other than the anchor instance as negatives~\cite{DBLP:conf/nips/YouCSCWS20,DBLP:journals/corr/abs-2106-04509}, we argue that, as with metric learning, contrastive representation learning benefits from hard negative samples (i.e., samples that are difficult to distinguish from the anchor sample). Hard negative instances can help guide the model to generate more discriminative representations. So we consider an additional hard negative mining scheme by treating molecules similar to the anchor instance as negatives. 

Instead of randomly choose graphs other than the anchor instance as negatives~\cite{DBLP:conf/nips/YouCSCWS20,DBLP:journals/corr/abs-2106-04509}, we consider an additional hard negative mining scheme by treating molecules similar to the anchor instance as negatives. Specifically, we represent each molecule by its Morgan Fingerprints~\cite{DBLP:journals/jcisd/RogersH10}, which perceive the presence of specific circular substructures around each atom in a molecule and encode it in a fixed length binary vector. Then we calculate the molecular similarity through their Tanimoto coefficient~\cite{DBLP:journals/jcheminf/BajuszRH15}:
\begin{small}
\begin{equation}
s(\boldsymbol{e}_1,\boldsymbol{e}_2) = \frac{N_{12}}{N_1+N_2-N_{12}},	
\end{equation}
\end{small}
where $\boldsymbol{e}_1,\boldsymbol{e}_2$ denotes the fingerprints, $N_1,N_2$ denotes the number of 1s in $\boldsymbol{e}_1,\boldsymbol{e}_2$ respectively, and $N_{12}$ denotes the number of 1s in the intersection of $\boldsymbol{e}_1,\boldsymbol{e}_2$. In order to ensure all molecules have negative samples, instead of setting a fixed threshold, we sorted samples by similarity and selected a batch of most similar molecules as the negative samples. 

\paragraph{Contrastive Loss.}
We augmented a minibatch of $N$ similar molecular graphs with knowledge-guided graph augmentation, resulting in a total of $2N$ graphs. Following ~\cite{DBLP:conf/nips/YouCSCWS20,DBLP:conf/icml/ChenK0H20}, given a positive pair, we treat the other $2(N-1)$ graphs within the same minibatch as hard negative samples. We utilize NT-Xent as our objective function like in~\cite{DBLP:conf/iclr/HjelmFLGBTB19,DBLP:conf/icml/ChenK0H20,DBLP:conf/nips/YouCSCWS20,DBLP:conf/isbi/CarseCM21}. The training objective for $(\mathcal{G}_i,\mathcal{G}_i')$ is defined as
    \begin{equation}
    \ell_{i}=-\log \frac{e^{ \rm{sim} \left(\boldsymbol{z}_i, \boldsymbol{z}_i'\right) / \tau}}{\sum_{j=1}^{N} \left( e^{ \rm{sim} \left(\boldsymbol{z}_i, \boldsymbol{z}_j'\right) / \tau} + e^{\rm{sim} \left(\boldsymbol{z}_i', \boldsymbol{z}_j\right) / \tau} \right)}
    ,
    \end{equation}
where $\tau$ denotes the temperature parameter and sim($\boldsymbol{z}_1, \boldsymbol{z}_2$) is the cosine similarity $\frac{\boldsymbol{z}_{1}^{\top} \boldsymbol{z}_{2}}{\left\|\boldsymbol{z}_{1}\right\| \cdot\left\|\boldsymbol{z}_{2}\right\|}$. The final loss is computed across all positive pairs in the minibatch.

\section{Experiments}
In this section, we conduct extensive experiments to examine the proposed method by answering the following questions:

\textbf{Q1}: How does KCL perform compared with state-of-the-art methods for molecular property prediction?
  
\textbf{Q2}: Does the knowledge-guided graph augmentation in Module 1 learns better representations than general augmentations?

\textbf{Q3}: How do knowledge feature initialization and graph encoders in Module 2 affect KCL?

\textbf{Q4}: How useful are the self-supervised contrastive learning and hard negative strategy in Module 3?

\textbf{Q5}: How can we interpret KCL(KMPNN) from a domain-specific perspective?

\begin{table*}[!h]
\small
  \centering
  \begin{tabular}{c|cccccc|cc}
	\hline
	\hline
    Task & \multicolumn{6}{c|}{Classification (ROC-AUC)} & \multicolumn{2}{c}{Regression (RMSE)} \\
%    & \multicolumn{6}{c|}{(Higher is better)} & \multicolumn{2}{c}{(Lower is better)} \\
	\hline
    Dataset & B\small{BBP} & T\small{ox21} & T\small{oxCast} & S\small{IDER} & C\small{linTox} & B\small{ACE} & E\small{SOL} & F\small{reeSolv} \\
    % \cline{2-9} 
    \#Molecules & 2039 & 7831 & 8575 & 1427 & 1478 & 1513 & 1128 & 642 \\
    % \cline{2-9}
    \#Tasks & 1 & 12 & 617 & 27 & 2 & 1 & 1 & 1 \\
	\hline
	G\small{CN~\cite{DBLP:conf/iclr/KipfW17}}    & 0.877 & 0.772  & 0.650   & 0.638  & 0.807   & 0.854  & 1.068 & 2.900  \\
%	\cline{2-9}
	W\small{eave~\cite{DBLP:journals/jcamd/KearnesMBPR16}}     & 0.837 & 0.741  & 0.678   & 0.621  & 0.823   & 0.791  & 1.158 & 2.398 \\
%	\cline{2-9}
	M\small{PNN~\cite{DBLP:conf/icml/GilmerSRVD17}}     & 0.913 & 0.808  & 0.691  & 0.641  & 0.879   & 0.815  & 1.167 & 2.185  \\
%	\cline{2-9}
	D\small{MPNN~\cite{DBLP:journals/jcisd/YangSJCEGGHKMPS19}}   & 0.919 & 0.826  & 0.718  & 0.632  & 0.897   & 0.852  & 0.980 & 2.177  \\
%	\cline{2-9}
	C\small{MPNN~\cite{DBLP:conf/ijcai/SongZNFLY20}}   & 0.927 & 0.806  & 0.738  & 0.636  & 0.902   & 0.869  & 0.798 & \underline{0.956}  \\
%	\cline{2-9}
	C\small{oMPT~\cite{DBLP:journals/corr/abs-2107-08773}}   & 0.938 & 0.809  & \underline{0.740}  & 0.634  & 0.934   & 0.871  & \underline{0.774} & 1.855  \\
	\hline
	 N\small{-GRAM~\cite{DBLP:conf/nips/LiuDL19}}  & 0.912 & 0.769 & -  & 0.632 & 0.870 & 0.876  & 1.100  & 2.512   \\
%	\cline{2-9}
	 Hu \small{et al.~\cite{DBLP:conf/iclr/HuLGZLPL20}}   & 0.915 & 0.811 & 0.714 & 0.614 & 0.762  & 0.851 & -           & -           \\
%	\cline{2-9}
	 G\small{ROVER~\cite{DBLP:conf/nips/RongBXX0HH20}} & \underline{0.940} &\underline{0.831} & 0.737 & \underline{0.658} & \underline{0.944} & \underline{0.894} & 0.831   & 1.544    \\
	\hline
	  K\small{CL(GCN)}       & 0.956 &  0.856 & \textbf{0.757}  &  0.666 & 0.945  &  \textbf{0.934} & \textbf{0.582} & 0.854   \\
	 K\small{CL(KMPNN)}  & \textbf{0.961} & \textbf{0.859} &  0.740  &  \textbf{0.671} & \textbf{0.958}  &  0.924 & 0.732 & \textbf{0.795}   \\
	\hline
	\hline
  \end{tabular}
   \caption{The property prediction performance (lower is better for regression) of KCL under the fine-tune protocol, compared with supervised learning (first group) and pre-training methods (second group) baselines on 8 datasets.}
    \label{fine-tune}
\end{table*}

\subsection{Experimental Setup}
\paragraph{Pre-training Data Collection.}
We collect 250K unlabeled molecules sampled from the ZINC15 datasets~\cite{DBLP:journals/jcisd/SterlingI15} to pre-train KCL. 
% We randomly split 10\% of unlabelled molecules as the validation sets for model selection and hyper-parameter tuning. 
\paragraph{Fine-tuning Tasks and Datasets.}
We use 8 benchmark datasets from the MoleculeNet~\cite{wu2018moleculenet} to perform the experiments, which cover a wide range of molecular tasks such as quantum mechanics, physical chemistry, biophysics, and physiology. For each dataset, as suggested by~\cite{wu2018moleculenet}, we apply three independent runs on three random-seeded random splitting or scaffold splitting with a ratio for train/validation/test as 8:1:1. Details of datasets and dataset splitting are deferred to Appendix~\ref{Dataset Description}. 
\paragraph{Baselines.}
We adopt three types of baselines:
\begin{itemize}
\item \textit{Supervised learning methods}: GCN~\cite{DBLP:conf/iclr/KipfW17} and Weave~\cite{DBLP:journals/jcamd/KearnesMBPR16} are two types of graph convolutional methods. MPNN~\cite{DBLP:conf/icml/GilmerSRVD17} and its variants DMPNN~\cite{DBLP:journals/jcisd/YangSJCEGGHKMPS19}, CMPNN~\cite{DBLP:conf/ijcai/SongZNFLY20}, CoMPT~\cite{DBLP:journals/corr/abs-2107-08773} consider the edge features and strengthen the message interactions between bonds and atoms during message passing.
\item \textit{Pre-trained methods}: N-GRAM~\cite{DBLP:conf/nips/LiuDL19} conducts node embeddings by predicting the node attributes. Hu et al.~\cite{DBLP:conf/iclr/HuLGZLPL20} and GROVER~\cite{DBLP:conf/nips/RongBXX0HH20} are pre-trained models incorporating both node-level and graph-level pretext tasks.
\item \textit{Graph contrastive learning baselines}: InfoGraph~\cite{DBLP:conf/iclr/SunHV020} maximizes the mutual information between nodes and graphs. MICRO-Graph~\cite{DBLP:conf/aaai/Subramonian21} is a motif-based contrastive method. GraphCL~\cite{DBLP:conf/nips/YouCSCWS20} constructs contrastive views of graph data via hand-picking ad-hoc augmentations. JOAO~\cite{DBLP:conf/icml/YouCSW21} automates the augmentation selection. MoCL~\cite{DBLP:journals/corr/abs-2106-04509} utilizes domain knowledge at two levels to assist representation learning.
\end{itemize}
\paragraph{Evaluation Protocol.}
The evaluation process follows two steps. We first pre-train the model and then evaluate the learned model on downstream tasks under two protocols.
\begin{itemize}
\item \textit{Fine-tune protocol}: To achieve the full potential of our model, given graph embeddings output by the KCL encoder, we use an additional MLP to predict the property of the molecule. Fine-tune parameters in the encoders and the MLP.
\item \textit{Linear Protocol}: For comparison of our model and contrastive learning baselines, we fix the graph embeddings from the pre-trained model, and train a linear classifier.
\end{itemize}	
\paragraph{Implementation details.}
We use the Adam optimizer with an initial learning rate of 0.0001 and batch size of 256. For pre-training models, the running epoch is fixed to 20. The temperature $\tau$ is set as 0.1. For downstream tasks, we use early stopping on the validation set. We apply the random search to obtain the best hyper-parameters based on the validation set. Our model is implemented with PyTorch~\cite{NEURIPS2019_9015} and Deep Graph Library~\cite{wang2019dgl}. We develop all codes on a Ubuntu Server with 4 GPUs (NVIDIA GeForce 1080Ti). More experimental details are available in Appendix~\ref{Implementation and Pre-training Details} and~\ref{Downstream Details}. 
% \add{can further compress this part to save space.}

\subsection{Performance Comparison (Q1 \& Q2)}
\paragraph{Performance under Fine-tune Protocol.}
We first examine whether the proposed KCL performs better than SOTA methods. Table~\ref{fine-tune} displays the complete results of supervised learning baselines and pre-trained methods, where the underlined cells indicate the previous SOTAs, and the cells with bold show the best result achieved by KCL. The Tox21, SIDER, and ClinTox are all multiple-task learning tasks, including totally 42 classification tasks. We also implemented two versions of our KCL model, the original molecular graph with GCN encoder and the augmented molecular graph with KMPNN as the encoder. 

Table~\ref{fine-tune} offers the following observations: {(1)} KCL consistently achieves the best performance on all datasets with large margins. The overall relative improvement is 7.1\% on all datasets (2.6\% on classification tasks and 20.4\% on regression tasks)\footnote{We use relative improvement to provide the unified descriptions.}. This notable performance improvement suggests the effectiveness of KCL for molecular property prediction tasks. {(2)} In the small dataset FreeSolv with only 642 labeled molecules, KCL gains a 16.8\% improvement over SOTA baselines. This confirms the strength of KCL since it can significantly help with tasks with very limited label information.

\paragraph{Performance under Linear Protocol.}

\begin{table}%[!h]
\small
  \centering
  \begin{tabular}{p{1.3cm}<{\centering}|p{0.6cm}<{\centering}p{0.58cm}<{\centering}p{0.75cm}<{\centering}p{0.6cm}<{\centering}p{0.65cm}<{\centering}p{0.65cm}<{\centering}}%	\hline
%  \begin{tabular}{c|cccccc}
%    & \multicolumn{6}{c|}{Classification} \\
	\hline
	\hline
    Dataset & B\small{BBP} & T\small{ox21} & T\small{oxCast} & S\small{IDER} & C\small{linTox} & B\small{ACE}  \\
	\hline
	 Node            & 0.843 & 0.728  & 0.633  & 0.577  & 0.635   & 0.746  \\
%	\cline{2-7}
	 Edge    & 0.833 & 0.715  & 0.619  & 0.605  & 0.630   & 0.657 \\
%	\cline{2-7}
	 Subgraph   & 0.815 & 0.727  & 0.625  & 0.583  & 0.603   & 0.629  \\
%	\cline{2-7}
	 Attribute     & 0.826 & 0.726  & 0.623  & 0.621  & 0.671 &  0.796   \\
	\hline
	I\small{nfoGraph} & 0.611 & 0.615  & 0.562   & 0.502  & 0.458   & 0.594    \\
	M\small{ICRO} & 0.830 & 0.718  & 0.595   & 0.573  & 0.735   & 0.708  \\
	G\small{raphCL} & 0.697 & 0.739  & 0.624   & 0.605  & 0.760   & 0.755  \\
	J\small{OAO} & 0.714 & 0.750  & 0.632   & 0.605  & \underline{0.813}   & 0.773  \\
	M\small{oCL} & \underline{0.905} & \underline{0.768}  & \underline{0.653}  & \underline{0.628}  & 0.750   & \underline{0.845}   \\
	\hline
	K\small{CL(G)} &\textbf{0.929} &  0.821 &  0.696  &  0.620 &\textbf{0.909} & \textbf{0.902} \\
	K\small{CL(K)}& 0.927 &  \textbf{0.825} &  \textbf{0.709}  &  \textbf{0.659} & 0.898 & 0.860 \\
	\hline
	\hline
	\end{tabular}
	\caption{The performance of KCL under the linear protocol on 6 datasets, compared with contrastive learning baselines. The metric is ROC-AUC.}
      \label{linear}
\end{table}

We next study whether the knowledge-guided graph augmentation in Module 1 helps learn better molecular representations. Table~\ref{linear} shows the comparison results of different augmentation (node dropping, edge perturbation, subgraph extraction and attribute masking) and contrastive learning methods. To be consistent with prior works and make the comparisons fair, we use the linear protocol, which is exactly what baselines have done, to evaluate the performance on classification datasets. Results on regression tasks are deferred to Appendix~\ref{Regression Results under Linear Protocol}.

Both versions of KCL produce better results compared to alternative graph augmentation methods (the first group in Table 3). This verifies our assumption that knowledge-guided graph augmentation does not violate the biological semantic in molecules and thus works better than other augmentations. Moreover, KCL gains a 7.0\% improvement over the previous best contrastive learning methods (the second group), which confirms that better representations of molecular graphs could be obtained by incorporating fundamental chemical domain knowledge and capturing microscopic associations between atoms.

\subsection{Ablation Study (Q3 \& Q4)}
 \begin{figure}[!t]
 \centering
 \includegraphics[width=0.95\columnwidth]{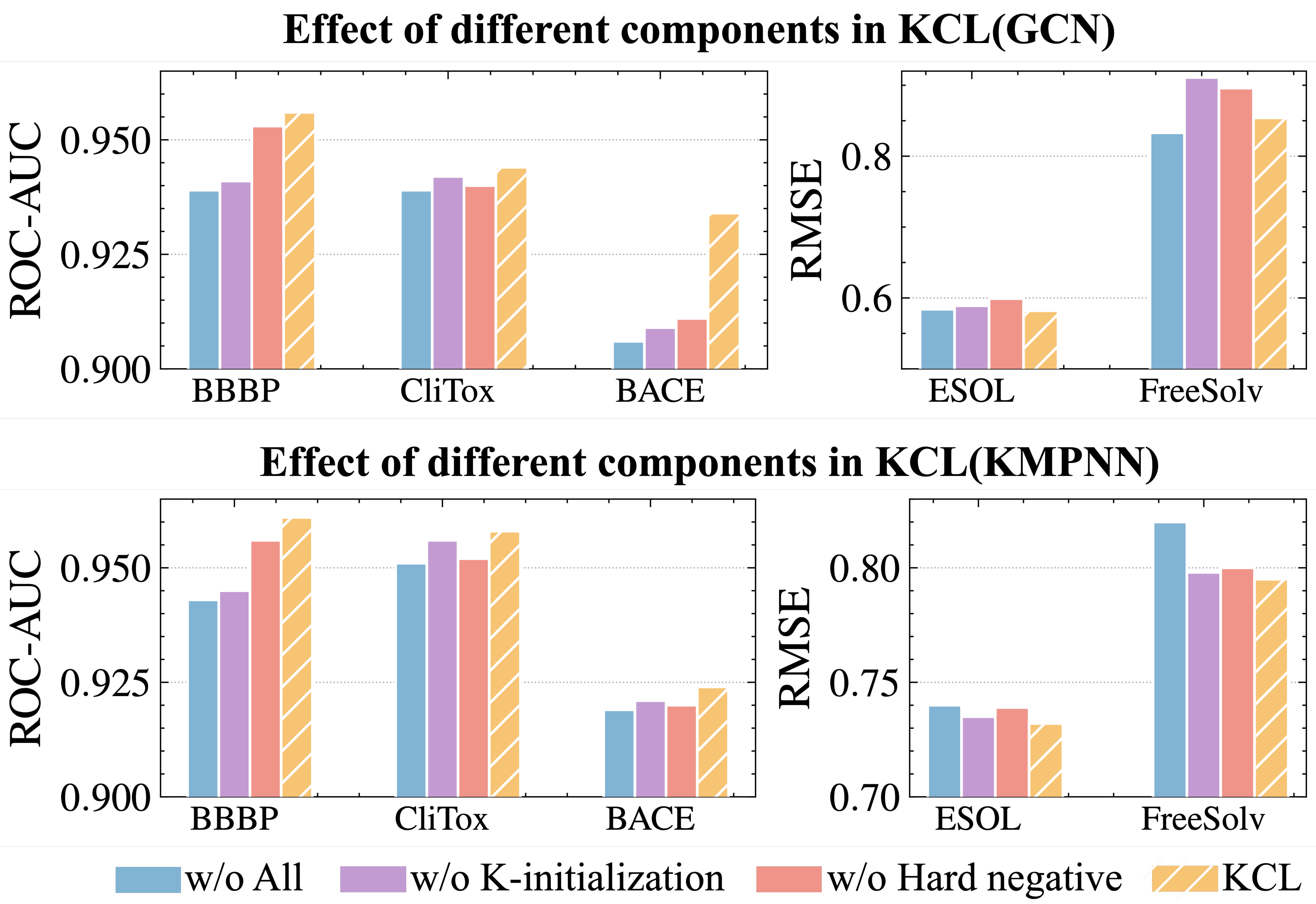} % Reduce the figure size so that it is slightly narrower than the column. Don't use precise values for figure width.This setup will avoid overfull boxes.
 \caption{Performance of KCL with different settings under the fine-tune protocol (lower is better for regression).}
 \label{ablation}
 \end{figure}

We then conducted ablation studies to investigate components in Module 1 and 2 that influence the performance of the proposed KCL framework. 

As shown in Figure~\ref{ablation}, KCL with knowledge feature initialization and hard negative mining scheme (bar in yellow) shows the best performance among all architectures. Models with random initialization and random negative sampling denoted by ``w/o ALL" almost always perform the worst. Excluding any of these two components can easily result in a decrease in performance. This illustrates that both knowledge feature initialization and hard negative mining strategy are necessary for KCL, because the former captures the structural triple information, while the latter guides the encoders to generate more discriminative representations.

\begin{table}%[!h]
\begin{small}
  \centering
   \begin{tabular}{ccc}
	\hline
	\hline
	Task & Classification & Regression \\
	\hline
	GCN(No contrast) & 0.766 & 1.984\\
    KMPNN(No contrast) & 0.806 & 1.531  \\
    \hline
    KCL(GIN)  & 0.849 & \underline{0.718} \\
    KCL(GAT)  & 0.850 & 0.724 \\
    KCL(GCN)  & \underline{0.852} & \underline{0.718} \\
    \hline
    KCL(RGCN)  & 0.831 & 1.008 \\
    KCL(MPNN)  & 0.833 &  0.927 \\
    KCL(KMPNN) & \underline{0.852} & \underline{0.765} \\
	\hline
	\hline
\end{tabular}
  \caption{Ablation results under the fine-tune protocol. Each value represents the average result of the task, and the underline marks the best in the group.}
 \label{tab:ablation}
 \end{small}
\end{table}

Since our graph encoders are pluggable, we replaced both GCN, KMPNN with other architectures to explore the impact of graph encoders.
The results in Table~\ref{tab:ablation} demonstrate that applying different encoders (e.g., GIN~\cite{DBLP:conf/iclr/XuHLJ19}, GAT~\cite{DBLP:conf/iclr/VelickovicCCRLB18}) on original molecular graphs has no significant impact on performance. In addition, we ignore the different types of nodes and edges in augmented graphs and replace KMPNN with previous heterogeneous graph neural network (RGCN~\cite{DBLP:conf/esws/SchlichtkrullKB18}) and general message passing framework (MPNN~\cite{DBLP:conf/icml/GilmerSRVD17}). The comparisons reveal that KMPNN has better expressive power by providing heterogeneous attentive message passing for different types of knowledge on the augmented molecular graphs.
The specific values are deferred to Appendix~\ref{Effect of Different Settings} and \ref{Effect of Different Encoders}.
% In addition, compared with previous heterogeneous graph neural network (RGCN~\cite{DBLP:conf/esws/SchlichtkrullKB18}) and general message passing framework (MPNN~\cite{DBLP:conf/icml/GilmerSRVD17}), KMPNN has better expressive power by providing heterogeneous attentive message passing for different types of knowledge on the augmented molecular graphs.
% The specific values are deferred to Appendix~\ref{Effect of Different Settings} and \ref{Effect of Different Encoders}.

To investigate the contribution of the self-supervision strategy, we compare the performances between KCL with and without contrastive learning under the fine-tune protocol (the counterpart under linear protocol is deferred to Appendix~\ref{Effect of Contrastive Learning}). We report the comparison results in Table~\ref{tab:ablation}. The self-supervised contrastive learning leads to a performance boost with an average increase of 8.5\% on classification and 56.9\% on regression over the model without contrastive learning. This confirms that contrastive learning can learn better representations by narrowing the distance between the structural view and the knowledgeable view in the latent space, and enhance the prediction performance of downstream tasks.

\subsection{Chemical Interpretability Analysis (Q5)}
Finally, we explore the interpretability of our model by visualizing the attention of each edge in a molecule. Specifically, we extracted and normalized the atom's attention weights to their neighbors from the last layer of KCL(KMPNN). 
\begin{figure}[!t]
\centering
\includegraphics[width=0.95\columnwidth]{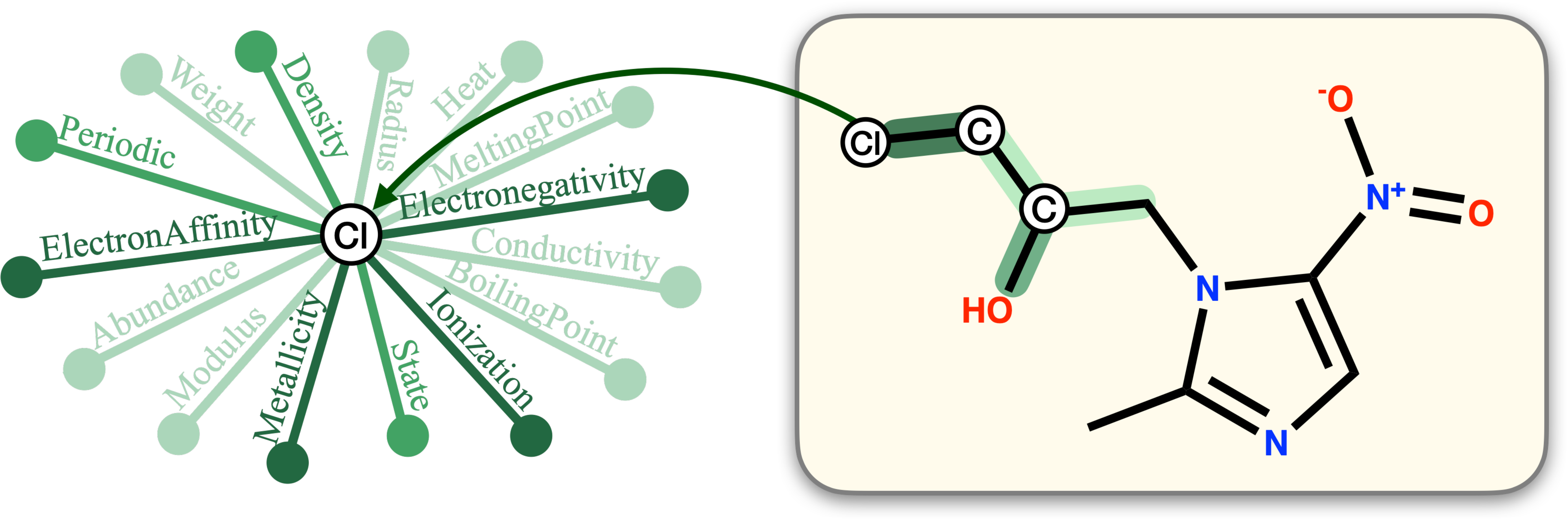} % Reduce the figure size so that it is slightly narrower than the column. Don't use precise values for figure width.This setup will avoid overfull boxes.
\caption{An attention visualization example of different types of neighbors (attributes and atoms) in the BBBP dataset. The attention weights assigned for bonds connected to the two C atoms are visualized on the right. The darker the color, the higher the attention.}
\label{attention}
\end{figure}

Figure~\ref{attention} illustrates an example in the BBBP dataset~\cite{DBLP:journals/jcisd/MartinsTPF12}. BBBP involves records of whether a compound carries the permeability property of penetrating the blood-brain barrier. As shown in the left part of the figure, atoms tend to assign more attention to their electron affinity, electronegativity, metallicity, and ionization. These attributes are closely related to atoms' ability to lose electrons. The strength of the atom's ability to gain or lose electrons will largely affect the polarity of the molecule, thereby affecting its permeability. In addition, more lively atomic neighbors are easier to be noticed, as illustrated on the right side of Figure~\ref{attention}. The element Cl has relatively higher electronegativity, so it has a stronger ability to obtain electrons. Also, the hydroxyl group promotes hydrophilicity and thus is assigned higher attention.
Another interesting observation is that fine-grained attributes (e.g., weight, radius) receive less attention than coarse-grained attributes (e.g., electron affinity, electronegativity, metallicity, and ionization). It is because coarse-grained attributes are more abstract and informative than fine-grained attributes, and therefore contain richer domain knowledge. This is in line with hierarchical machine learning where coarse-grained features at higher levels can be seen as a summary of fine-grained features in terms of target prediction.
%Attributes with a macroscopic degree in between (e.g., density, periodic) are given medium attention. 
%Since most of the phenomena that humans can easily observe are relatively macroscopic (e.g., permeability, toxicity), we claim that such attention distribution is intuitive. 
More examples and discussions on other datasets are in Appendix~\ref{Chemical Interpretability Analysis}.

\section{Conclusion and Future Work}
This paper aims to incorporate fundamental domain knowledge into molecular graph representation learning.
We construct Element KG to build microscopic connections between elements, and propose to utilize knowledge in the KCL framework to enhance molecular graph contrastive learning.
%(1)\textit{Knowledge-guided graph augmention module}: we , and use it to extend the graph augmentation. (2)\textit{Pluggable graph encoders}: we adopt a commonly used graph encoder to capture the strucural information of the original molecular graphs and introduce a KMPNN encoder to extract complex knowledge in the augmented graphs. (3)\textit{Hard contrastive objective}: we design a hard negative mining strategy and train the encoders to maximize the agreement between positive pairs and the discrepancy between hard negatives. 
We demonstrate the effectiveness of KCL under both fine-tune and linear protocols, and experiments show that KCL excels previous methods with better interpretation and representation capability.

In the future, we intend to extend our work in several aspects. First, we would introduce different granularity of domain knowledge to enrich Chemical Element KG. Also, we will improve the current KG with more description logics defined in OWL2, such as more object properties and axioms. 
% Also, we would explore if Element KG can also boost other categories of tasks in chemical and biological domains. 
Third, we will open-source Chemical Element KG, continue to improve its quality and expand its scale. 
% it is interesting to develop more self-supervision pretext tasks with the guidance of knowledge in the Element KG.
 
\newpage

\section*{Acknowledgement}
We want to express gratitude to the anonymous reviewers for their hard work and kind comments. This work is funded by NSFCU19B2027/91846204, national key research program 2018YFB1402800.

% Use \bibliography{yourbibfile} instead or the References section will not appear in your paper
\bibliography{aaai22}

\begin{thebibliography}{50}
\providecommand{\natexlab}[1]{#1}

\bibitem[{tox(2017)}]{tox21}
 2017.
\newblock Tox21 challenge.
\newblock \emph{https://tripod.nih.gov/tox21/challenge/}.

\bibitem[{Bajusz, R{\'{a}}cz, and
  H{\'{e}}berger(2015)}]{DBLP:journals/jcheminf/BajuszRH15}
Bajusz, D.; R{\'{a}}cz, A.; and H{\'{e}}berger, K. 2015.
\newblock Why is Tanimoto index an appropriate choice for fingerprint-based
  similarity calculations?
\newblock \emph{J. Cheminformatics}, 7: 20:1--20:13.

\bibitem[{Bemis and Murcko(1996)}]{bemis1996properties}
Bemis, G.~W.; and Murcko, M.~A. 1996.
\newblock The properties of known drugs. 1. Molecular frameworks.
\newblock \emph{Journal of medicinal chemistry}, 39(15): 2887--2893.

\bibitem[{Carse, Carey, and McKenna(2021)}]{DBLP:conf/isbi/CarseCM21}
Carse, J.; Carey, F.~A.; and McKenna, S.~J. 2021.
\newblock Unsupervised Representation Learning From Pathology Images With
  Multi-Directional Contrastive Predictive Coding.
\newblock In \emph{{ISBI}}, 1254--1258. {IEEE}.

\bibitem[{Chen et~al.(2021)Chen, Zheng, Song, Rao, and
  Yang}]{DBLP:journals/corr/abs-2107-08773}
Chen, J.; Zheng, S.; Song, Y.; Rao, J.; and Yang, Y. 2021.
\newblock Learning Attributed Graph Representations with Communicative Message
  Passing Transformer.
\newblock \emph{CoRR}, abs/2107.08773.

\bibitem[{Chen et~al.(2020)Chen, Kornblith, Norouzi, and
  Hinton}]{DBLP:conf/icml/ChenK0H20}
Chen, T.; Kornblith, S.; Norouzi, M.; and Hinton, G.~E. 2020.
\newblock A Simple Framework for Contrastive Learning of Visual
  Representations.
\newblock In \emph{{ICML}}, volume 119 of \emph{Proceedings of Machine Learning
  Research}, 1597--1607. {PMLR}.

\bibitem[{Cho et~al.(2014)Cho, van Merrienboer, Bahdanau, and
  Bengio}]{DBLP:conf/ssst/ChoMBB14}
Cho, K.; van Merrienboer, B.; Bahdanau, D.; and Bengio, Y. 2014.
\newblock On the Properties of Neural Machine Translation: Encoder-Decoder
  Approaches.
\newblock In \emph{SSST@EMNLP}, 103--111. Association for Computational
  Linguistics.

\bibitem[{Delaney(2004)}]{delaney2004esol}
Delaney, J.~S. 2004.
\newblock ESOL: estimating aqueous solubility directly from molecular
  structure.
\newblock \emph{Journal of chemical information and computer sciences}, 44(3):
  1000--1005.

\bibitem[{Delmas et~al.(2021)Delmas, Filangi, Paulhe, Vinson, Duperier,
  Garrier, Saunier, Pitarch, Jourdan, Giacomoni et~al.}]{delmas2021building}
Delmas, M.; Filangi, O.; Paulhe, N.; Vinson, F.; Duperier, C.; Garrier, W.;
  Saunier, P.-E.; Pitarch, Y.; Jourdan, F.; Giacomoni, F.; et~al. 2021.
\newblock Building a Knowledge Graph from public databases and scientific
  literature to extract associations between chemicals and diseases.
\newblock \emph{bioRxiv}.

\bibitem[{Duvenaud et~al.(2015)Duvenaud, Maclaurin, Aguilera{-}Iparraguirre,
  G{\'{o}}mez{-}Bombarelli, Hirzel, Aspuru{-}Guzik, and
  Adams}]{DBLP:conf/nips/DuvenaudMABHAA15}
Duvenaud, D.; Maclaurin, D.; Aguilera{-}Iparraguirre, J.;
  G{\'{o}}mez{-}Bombarelli, R.; Hirzel, T.; Aspuru{-}Guzik, A.; and Adams,
  R.~P. 2015.
\newblock Convolutional Networks on Graphs for Learning Molecular Fingerprints.
\newblock In \emph{{NIPS}}, 2224--2232.

\bibitem[{Fang et~al.(2021)Fang, Yang, Zhuang, Shao, Fan, and
  Chen}]{DBLP:journals/corr/abs-2103-13047}
Fang, Y.; Yang, H.; Zhuang, X.; Shao, X.; Fan, X.; and Chen, H. 2021.
\newblock Knowledge-aware Contrastive Molecular Graph Learning.
\newblock \emph{CoRR}, abs/2103.13047.

\bibitem[{Gayvert, Madhukar, and Elemento(2016)}]{gayvert2016data}
Gayvert, K.~M.; Madhukar, N.~S.; and Elemento, O. 2016.
\newblock A data-driven approach to predicting successes and failures of
  clinical trials.
\newblock \emph{Cell chemical biology}, 23(10): 1294--1301.

\bibitem[{Gilmer et~al.(2017)Gilmer, Schoenholz, Riley, Vinyals, and
  Dahl}]{DBLP:conf/icml/GilmerSRVD17}
Gilmer, J.; Schoenholz, S.~S.; Riley, P.~F.; Vinyals, O.; and Dahl, G.~E. 2017.
\newblock Neural Message Passing for Quantum Chemistry.
\newblock In \emph{{ICML}}, volume~70 of \emph{Proceedings of Machine Learning
  Research}, 1263--1272. {PMLR}.

\bibitem[{Hassani and Ahmadi(2020)}]{DBLP:conf/icml/HassaniA20}
Hassani, K.; and Ahmadi, A. H.~K. 2020.
\newblock Contrastive Multi-View Representation Learning on Graphs.
\newblock In \emph{{ICML}}, volume 119 of \emph{Proceedings of Machine Learning
  Research}, 4116--4126. {PMLR}.

\bibitem[{Hjelm et~al.(2019)Hjelm, Fedorov, Lavoie{-}Marchildon, Grewal,
  Bachman, Trischler, and Bengio}]{DBLP:conf/iclr/HjelmFLGBTB19}
Hjelm, R.~D.; Fedorov, A.; Lavoie{-}Marchildon, S.; Grewal, K.; Bachman, P.;
  Trischler, A.; and Bengio, Y. 2019.
\newblock Learning deep representations by mutual information estimation and
  maximization.
\newblock In \emph{{ICLR}}. OpenReview.net.

\bibitem[{Hu et~al.(2020)Hu, Liu, Gomes, Zitnik, Liang, Pande, and
  Leskovec}]{DBLP:conf/iclr/HuLGZLPL20}
Hu, W.; Liu, B.; Gomes, J.; Zitnik, M.; Liang, P.; Pande, V.~S.; and Leskovec,
  J. 2020.
\newblock Strategies for Pre-training Graph Neural Networks.
\newblock In \emph{{ICLR}}. OpenReview.net.

\bibitem[{Jastrzebski et~al.(2016)Jastrzebski, Lesniak, Czarnecki, and
  .}]{DBLP:journals/corr/JastrzebskiLC16}
Jastrzebski, S.; Lesniak, D.; Czarnecki, W.~M.; and . 2016.
\newblock Learning to {SMILE(S)}.
\newblock \emph{CoRR}, abs/1602.06289.

\bibitem[{Kearnes et~al.(2016)Kearnes, McCloskey, Berndl, Pande, and
  Riley}]{DBLP:journals/jcamd/KearnesMBPR16}
Kearnes, S.~M.; McCloskey, K.; Berndl, M.; Pande, V.~S.; and Riley, P. 2016.
\newblock Molecular graph convolutions: moving beyond fingerprints.
\newblock \emph{J. Comput. Aided Mol. Des.}, 30(8): 595--608.

\bibitem[{Kipf and Welling(2017)}]{DBLP:conf/iclr/KipfW17}
Kipf, T.~N.; and Welling, M. 2017.
\newblock Semi-Supervised Classification with Graph Convolutional Networks.
\newblock In \emph{{ICLR} (Poster)}. OpenReview.net.

\bibitem[{Kuhn et~al.(2016)Kuhn, Letunic, Jensen, and
  Bork}]{DBLP:journals/nar/KuhnLJB16}
Kuhn, M.; Letunic, I.; Jensen, L.~J.; and Bork, P. 2016.
\newblock The {SIDER} database of drugs and side effects.
\newblock \emph{Nucleic Acids Res.}, 44(Database-Issue): 1075--1079.

\bibitem[{Lin et~al.(2020)Lin, Quan, Wang, Ma, and
  Zeng}]{DBLP:conf/ijcai/LinQWMZ20}
Lin, X.; Quan, Z.; Wang, Z.; Ma, T.; and Zeng, X. 2020.
\newblock {KGNN:} Knowledge Graph Neural Network for Drug-Drug Interaction
  Prediction.
\newblock In \emph{{IJCAI}}, 2739--2745. ijcai.org.

\bibitem[{Liu et~al.(2019)Liu, Demirel, Liang, and .}]{DBLP:conf/nips/LiuDL19}
Liu, S.; Demirel, M.~F.; Liang, Y.; and . 2019.
\newblock N-Gram Graph: Simple Unsupervised Representation for Graphs, with
  Applications to Molecules.
\newblock In \emph{NeurIPS}, 8464--8476.

\bibitem[{Martins et~al.(2012)Martins, Teixeira, Pinheiro, and
  Falc{\~{a}}o}]{DBLP:journals/jcisd/MartinsTPF12}
Martins, I.~F.; Teixeira, A.~L.; Pinheiro, L.; and Falc{\~{a}}o, A.~O. 2012.
\newblock A Bayesian Approach to \emph{in Silico} Blood-Brain Barrier
  Penetration Modeling.
\newblock \emph{J. Chem. Inf. Model.}, 52(6): 1686--1697.

\bibitem[{Mobley and Guthrie(2014)}]{mobley2014freesolv}
Mobley, D.~L.; and Guthrie, J.~P. 2014.
\newblock FreeSolv: a database of experimental and calculated hydration free
  energies, with input files.
\newblock \emph{Journal of computer-aided molecular design}, 28(7): 711--720.

\bibitem[{Paszke et~al.(2019)Paszke, Gross, Massa, Lerer, Bradbury, Chanan,
  Killeen, Lin, Gimelshein, Antiga, Desmaison, Kopf, Yang, DeVito, Raison,
  Tejani, Chilamkurthy, Steiner, Fang, Bai, and Chintala}]{NEURIPS2019_9015}
Paszke, A.; Gross, S.; Massa, F.; Lerer, A.; Bradbury, J.; Chanan, G.; Killeen,
  T.; Lin, Z.; Gimelshein, N.; Antiga, L.; Desmaison, A.; Kopf, A.; Yang, E.;
  DeVito, Z.; Raison, M.; Tejani, A.; Chilamkurthy, S.; Steiner, B.; Fang, L.;
  Bai, J.; and Chintala, S. 2019.
\newblock PyTorch: An Imperative Style, High-Performance Deep Learning Library.
\newblock In Wallach, H.; Larochelle, H.; Beygelzimer, A.; d\textquotesingle
  Alch\'{e}-Buc, F.; Fox, E.; and Garnett, R., eds., \emph{Advances in Neural
  Information Processing Systems 32}, 8024--8035. Curran Associates, Inc.

\bibitem[{Richard et~al.(2016)Richard, Judson, Houck, Grulke, Volarath,
  Thillainadarajah, Yang, Rathman, Martin, Wambaugh
  et~al.}]{richard2016toxcast}
Richard, A.~M.; Judson, R.~S.; Houck, K.~A.; Grulke, C.~M.; Volarath, P.;
  Thillainadarajah, I.; Yang, C.; Rathman, J.; Martin, M.~T.; Wambaugh, J.~F.;
  et~al. 2016.
\newblock ToxCast chemical landscape: paving the road to 21st century
  toxicology.
\newblock \emph{Chemical research in toxicology}, 29(8): 1225--1251.

\bibitem[{Rogers and Hahn(2010)}]{DBLP:journals/jcisd/RogersH10}
Rogers, D.; and Hahn, M. 2010.
\newblock Extended-Connectivity Fingerprints.
\newblock \emph{J. Chem. Inf. Model.}, 50(5): 742--754.

\bibitem[{Rong et~al.(2020)Rong, Bian, Xu, Xie, Wei, Huang, and
  Huang}]{DBLP:conf/nips/RongBXX0HH20}
Rong, Y.; Bian, Y.; Xu, T.; Xie, W.; Wei, Y.; Huang, W.; and Huang, J. 2020.
\newblock Self-Supervised Graph Transformer on Large-Scale Molecular Data.
\newblock In \emph{NeurIPS}.

\bibitem[{Schlichtkrull et~al.(2018)Schlichtkrull, Kipf, Bloem, van~den Berg,
  Titov, and Welling}]{DBLP:conf/esws/SchlichtkrullKB18}
Schlichtkrull, M.~S.; Kipf, T.~N.; Bloem, P.; van~den Berg, R.; Titov, I.; and
  Welling, M. 2018.
\newblock Modeling Relational Data with Graph Convolutional Networks.
\newblock In \emph{{ESWC}}, volume 10843 of \emph{Lecture Notes in Computer
  Science}, 593--607. Springer.

\bibitem[{Shang et~al.(2019)Shang, Ma, Xiao, and
  Sun}]{DBLP:conf/ijcai/ShangMXS19}
Shang, J.; Ma, T.; Xiao, C.; and Sun, J. 2019.
\newblock Pre-training of Graph Augmented Transformers for Medication
  Recommendation.
\newblock In \emph{{IJCAI}}, 5953--5959. ijcai.org.

\bibitem[{Song et~al.(2020)Song, Zheng, Niu, Fu, Lu, and
  Yang}]{DBLP:conf/ijcai/SongZNFLY20}
Song, Y.; Zheng, S.; Niu, Z.; Fu, Z.; Lu, Y.; and Yang, Y. 2020.
\newblock Communicative Representation Learning on Attributed Molecular Graphs.
\newblock In \emph{{IJCAI}}, 2831--2838. ijcai.org.

\bibitem[{Sterling and Irwin(2015)}]{DBLP:journals/jcisd/SterlingI15}
Sterling, T.; and Irwin, J.~J. 2015.
\newblock {ZINC} 15 - Ligand Discovery for Everyone.
\newblock \emph{J. Chem. Inf. Model.}, 55(11): 2324--2337.

\bibitem[{Subramanian et~al.(2016)Subramanian, Ramsundar, Pande, and
  Denny}]{subramanian2016computational}
Subramanian, G.; Ramsundar, B.; Pande, V.; and Denny, R.~A. 2016.
\newblock Computational modeling of $\beta$-secretase 1 (BACE-1) inhibitors
  using ligand based approaches.
\newblock \emph{Journal of chemical information and modeling}, 56(10):
  1936--1949.

\bibitem[{Subramonian(2021)}]{DBLP:conf/aaai/Subramonian21}
Subramonian, A. 2021.
\newblock MOTIF-Driven Contrastive Learning of Graph Representations.
\newblock In \emph{{AAAI}}, 15980--15981. {AAAI} Press.

\bibitem[{Sun et~al.(2020)Sun, Hoffmann, Verma, and
  Tang}]{DBLP:conf/iclr/SunHV020}
Sun, F.; Hoffmann, J.; Verma, V.; and Tang, J. 2020.
\newblock InfoGraph: Unsupervised and Semi-supervised Graph-Level
  Representation Learning via Mutual Information Maximization.
\newblock In \emph{{ICLR}}. OpenReview.net.

\bibitem[{Sun, Lin, and Zhu(2020)}]{DBLP:conf/aaai/SunLZ20}
Sun, K.; Lin, Z.; and Zhu, Z. 2020.
\newblock Multi-Stage Self-Supervised Learning for Graph Convolutional Networks
  on Graphs with Few Labeled Nodes.
\newblock In \emph{{AAAI}}, 5892--5899. {AAAI} Press.

\bibitem[{Sun et~al.(2021)Sun, Xing, Wang, Chen, and
  Zhou}]{DBLP:journals/corr/abs-2106-04509}
Sun, M.; Xing, J.; Wang, H.; Chen, B.; and Zhou, J. 2021.
\newblock MoCL: Contrastive Learning on Molecular Graphs with Multi-level
  Domain Knowledge.
\newblock \emph{CoRR}, abs/2106.04509.

\bibitem[{Sun et~al.(2019)Sun, Deng, Nie, and Tang}]{DBLP:conf/iclr/SunDNT19}
Sun, Z.; Deng, Z.; Nie, J.; and Tang, J. 2019.
\newblock RotatE: Knowledge Graph Embedding by Relational Rotation in Complex
  Space.
\newblock In \emph{{ICLR} (Poster)}. OpenReview.net.

\bibitem[{Velickovic et~al.(2018)Velickovic, Cucurull, Casanova, Romero,
  Li{\`{o}}, and Bengio}]{DBLP:conf/iclr/VelickovicCCRLB18}
Velickovic, P.; Cucurull, G.; Casanova, A.; Romero, A.; Li{\`{o}}, P.; and
  Bengio, Y. 2018.
\newblock Graph Attention Networks.
\newblock In \emph{{ICLR} (Poster)}. OpenReview.net.

\bibitem[{Velickovic et~al.(2019)Velickovic, Fedus, Hamilton, Li{\`{o}},
  Bengio, and Hjelm}]{DBLP:conf/iclr/VelickovicFHLBH19}
Velickovic, P.; Fedus, W.; Hamilton, W.~L.; Li{\`{o}}, P.; Bengio, Y.; and
  Hjelm, R.~D. 2019.
\newblock Deep Graph Infomax.
\newblock In \emph{{ICLR} (Poster)}. OpenReview.net.

\bibitem[{Vinyals, Bengio, and Kudlur(2016)}]{DBLP:journals/corr/VinyalsBK15}
Vinyals, O.; Bengio, S.; and Kudlur, M. 2016.
\newblock Order Matters: Sequence to sequence for sets.
\newblock In \emph{{ICLR} (Poster)}.

\bibitem[{Wang et~al.(2019)Wang, Zheng, Ye, Gan, Li, Song, Zhou, Ma, Yu, Gai,
  Xiao, He, Karypis, Li, and Zhang}]{wang2019dgl}
Wang, M.; Zheng, D.; Ye, Z.; Gan, Q.; Li, M.; Song, X.; Zhou, J.; Ma, C.; Yu,
  L.; Gai, Y.; Xiao, T.; He, T.; Karypis, G.; Li, J.; and Zhang, Z. 2019.
\newblock Deep Graph Library: A Graph-Centric, Highly-Performant Package for
  Graph Neural Networks.
\newblock \emph{arXiv preprint arXiv:1909.01315}.

\bibitem[{Wu et~al.(2018{\natexlab{a}})Wu, Ramsundar, Feinberg, Gomes,
  Geniesse, Pappu, Leswing, and Pande}]{wu2018moleculenet}
Wu, Z.; Ramsundar, B.; Feinberg, E.~N.; Gomes, J.; Geniesse, C.; Pappu, A.~S.;
  Leswing, K.; and Pande, V. 2018{\natexlab{a}}.
\newblock MoleculeNet: a benchmark for molecular machine learning.
\newblock \emph{Chemical science}, 9(2): 513--530.

\bibitem[{Wu et~al.(2018{\natexlab{b}})Wu, Xiong, Yu, and
  Lin}]{DBLP:conf/cvpr/WuXYL18}
Wu, Z.; Xiong, Y.; Yu, S.~X.; and Lin, D. 2018{\natexlab{b}}.
\newblock Unsupervised Feature Learning via Non-Parametric Instance
  Discrimination.
\newblock In \emph{{CVPR}}, 3733--3742. {IEEE} Computer Society.

\bibitem[{Xu et~al.(2019)Xu, Hu, Leskovec, and
  Jegelka}]{DBLP:conf/iclr/XuHLJ19}
Xu, K.; Hu, W.; Leskovec, J.; and Jegelka, S. 2019.
\newblock How Powerful are Graph Neural Networks?
\newblock In \emph{{ICLR}}. OpenReview.net.

\bibitem[{Xu et~al.(2017)Xu, Wang, Zhu, and Huang}]{DBLP:conf/bcb/0005WZH17}
Xu, Z.; Wang, S.; Zhu, F.; and Huang, J. 2017.
\newblock Seq2seq Fingerprint: An Unsupervised Deep Molecular Embedding for
  Drug Discovery.
\newblock In \emph{{BCB}}, 285--294. {ACM}.

\bibitem[{Yang et~al.(2019)Yang, Swanson, Jin, Coley, Eiden, Gao,
  Guzman{-}Perez, Hopper, Kelley, Mathea, Palmer, Settels, Jaakkola, Jensen,
  and Barzilay}]{DBLP:journals/jcisd/YangSJCEGGHKMPS19}
Yang, K.; Swanson, K.; Jin, W.; Coley, C.~W.; Eiden, P.; Gao, H.;
  Guzman{-}Perez, A.; Hopper, T.; Kelley, B.; Mathea, M.; Palmer, A.; Settels,
  V.; Jaakkola, T.~S.; Jensen, K.~F.; and Barzilay, R. 2019.
\newblock Analyzing Learned Molecular Representations for Property Prediction.
\newblock \emph{J. Chem. Inf. Model.}, 59(8): 3370--3388.

\bibitem[{Yasunaga and Liang(2020)}]{DBLP:conf/icml/YasunagaL20}
Yasunaga, M.; and Liang, P. 2020.
\newblock Graph-based, Self-Supervised Program Repair from Diagnostic Feedback.
\newblock In \emph{{ICML}}, volume 119 of \emph{Proceedings of Machine Learning
  Research}, 10799--10808. {PMLR}.

\bibitem[{You et~al.(2021)You, Chen, Shen, and Wang}]{DBLP:conf/icml/YouCSW21}
You, Y.; Chen, T.; Shen, Y.; and Wang, Z. 2021.
\newblock Graph Contrastive Learning Automated.
\newblock In \emph{{ICML}}, volume 139 of \emph{Proceedings of Machine Learning
  Research}, 12121--12132. {PMLR}.

\bibitem[{You et~al.(2020)You, Chen, Sui, Chen, Wang, and
  Shen}]{DBLP:conf/nips/YouCSCWS20}
You, Y.; Chen, T.; Sui, Y.; Chen, T.; Wang, Z.; and Shen, Y. 2020.
\newblock Graph Contrastive Learning with Augmentations.
\newblock In \emph{NeurIPS}.

\end{thebibliography}

\newpage
\appendix
\section*{Appendix}
% \section{rotate}
%  The RotatE method defines each relation as a rotation from the source entity to the target entity, expecting that $\boldsymbol{t}=\boldsymbol{h} \circ \boldsymbol{r}$, where $\boldsymbol{h},\boldsymbol{r},\boldsymbol{t}$ are the embeddings, the modulus $|\boldsymbol{r}|=1$ and $\circ$ denotes the Hadamard product. \add{can add more details and insights}
 
\section{Details of KCL}
\subsection{Feature Initialization}\label{Feature Initialization}
The feature initialization contains two parts: 1) Atom / bond feature initialization. We randomly initialize different vectors for each type of atoms and bonds, with dimensions 128 and 64 respectively. We try a variety of different combinations of dimensions, and find that the impact on performance was not significant. Therefore, here we select the dimensions when the performance is optimal.
2) Attribute / relation feature initialization. We use RotatE~\cite{DBLP:conf/iclr/SunDNT19} which defines each relation as a rotation in the complex vector space to train Chemical Element KG. Its score function is formulated as follows:
\begin{small}
\begin{equation}
f(\mathbf{h}, \mathbf{r}, \mathbf{t})=\|\mathbf{h} \circ \mathbf{r}-\mathbf{t}\|,
\end{equation}
\end{small}
where $\mathbf{h}$, $\mathbf{r}$, $\mathbf{t}$ denote the embedding of head, relation and tail respectively, and $\circ$ is the Hadmard product. The dimensions of attributes and relations are also 128 and 64.

\subsection{Architecture of KMPNN}\label{appendix:architecture}
 \begin{figure}[!h]
\centering
\includegraphics[width=0.8\columnwidth]{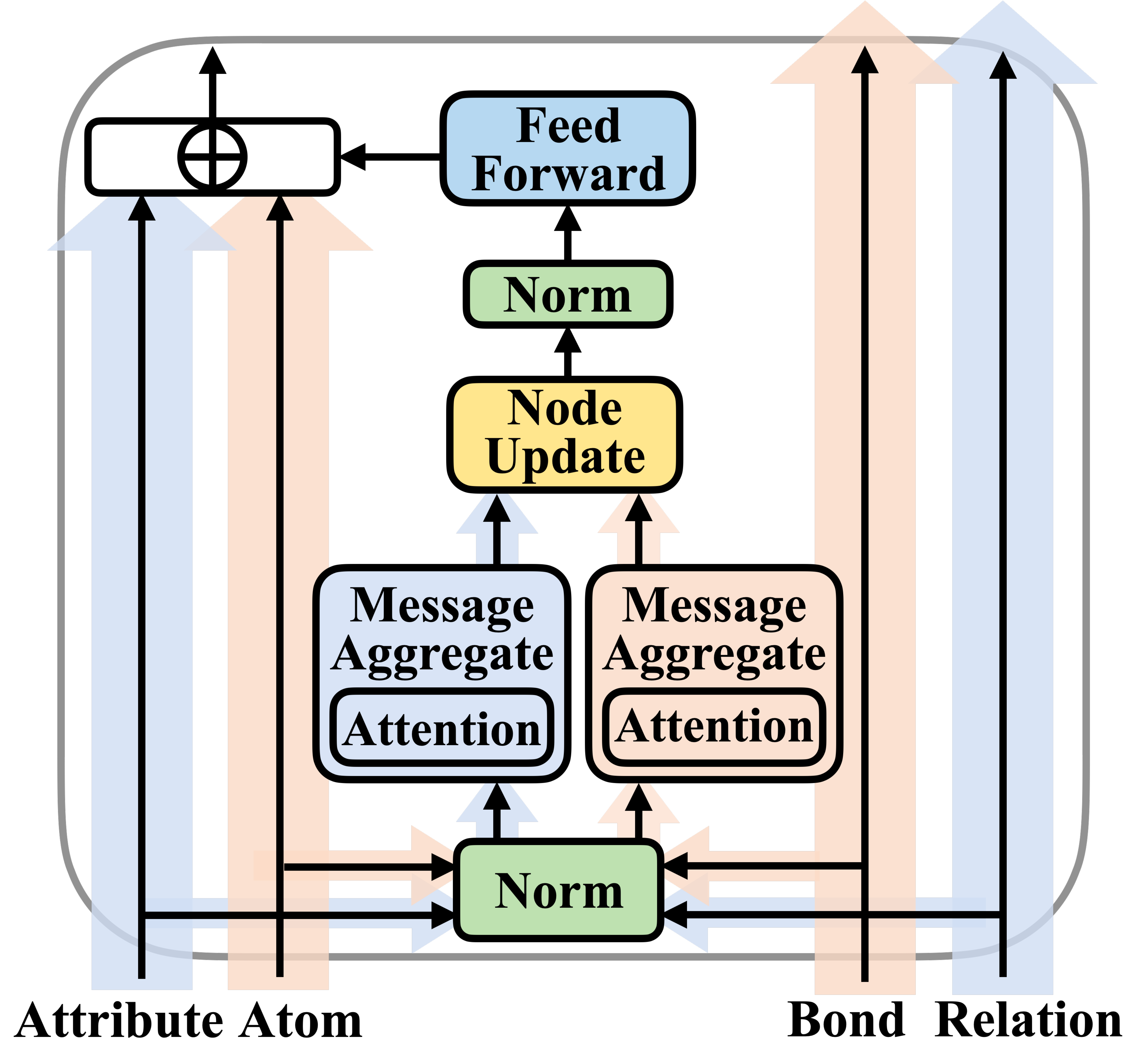} % Reduce the figure size so that it is slightly narrower than the column. Don't use precise values for figure width.This setup will avoid overfull boxes.
\caption{Architecture of KMPNN.}
\label{architecture}
\end{figure}
Figure~\ref{architecture} illustrates the architecture of KMPNN. We applied heterogeneous message passing (arrows in different colors) to neighbors representing atoms and attributes in the augmented molecular graph, respectively. In this way, the node information corresponding to the same type of knowledge share parameters during message passing. Moreover, we extend the message passing process by the self-attention mechanism. We use different parameters to compute attention coefficients between atoms and attributes, atoms and atoms. This attentive message passing allows assigning different attention values to neighbor nodes. 
% Then we describe the details of message diffusion and global pooling. In message diffusion module, we collect the messages from their neighboring edges in message aggregation, 
% \begin{small}
% \begin{equation}
% \boldsymbol{m}^k(v)= \sum_{k \in \mathcal{N}_{u}} \mathrm{M}\scriptstyle\mathrm{SG}\displaystyle(\boldsymbol{h}^{k-1}(e_{uv}) , \boldsymbol{h}^{k-1}(u)),
% \end{equation}
% \end{small}
% and apply GRU as the update function,
% \begin{small}
% \begin{equation}
% \boldsymbol{h}^{k}(v) =\mathrm{GRU}(\boldsymbol{h}^{k-1}(v), \boldsymbol{m}^k(v)),
% \end{equation}
% \end{small}
% where GRU is the Gated Recurrent Unit introduced in~\cite{DBLP:conf/ssst/ChoMBB14}. After $K$ steps' iteration, a readout operator is applied to get a graph-level representation for the molecule. 
% We adopt a set2set for global pooling following ~\cite{DBLP:journals/corr/VinyalsBK15},
% \begin{small} 
% \begin{equation}
% \boldsymbol{z} = \mathrm{Set2set}(\boldsymbol{h}^{k-1}(v)),
% \end{equation}
% \end{small}
% which is specifically designed to operate on sets and should have more expressive power than simply summing the final node states.

\subsection{GNN-based Encoder}
As for the original molecular graph $G$, we apply a Graph Neural Network (GNN)-based encoder to extract its graph-level representation. 
Formally, the message passing operation in iteration $k$ can be formulated as:
\begin{small}
\begin{equation}
\boldsymbol{h}^{k}(v) =\mathrm{U}(\boldsymbol{h}^{k-1}(v),\mathrm{A}\scriptstyle\mathrm{GG}\displaystyle(\{ \boldsymbol{h}^{k-1}(u)), \forall u \in \mathcal{N}_{v} \})),
\end{equation}
\end{small}
there are several ways of choosing $\mathrm{A}\scriptstyle\mathrm{GG}$, such as mean, sum, and max pooling. Then we convert them to graph embedding with a readout function. Note that no constraints are imposed on the GNN architecture, and experiments have also proved that different architectures have little impact on performance. So here we adopt the most classic GCN~\cite{DBLP:conf/iclr/KipfW17} architecture, combined with weighted sum and max pooling to compute the final readout.

% \subsection{Projection Head}
% A non-linear transformation named the projection head maps both original and augmented representations to the space where contrastive loss is applied, as advocated in~\cite{DBLP:conf/icml/ChenK0H20}. By leveraging the projection head, more information can be formed and maintained in representations. A perceptron (MLP) with one hidden layer act as the projection head in our model. After pre-training is completed, we throw away the projection head and only use the encoder for downstream tasks.

\section{Details about Experimental Setup}

\begin{table*}%[!h]
    \centering
\begin{tabular}{c|c|c|c|c|c|c}
\hline
\hline
\multicolumn{1}{c|}{\textbf{Type}} & \multicolumn{1}{c|}{\textbf{Category}} & \textbf{Dataset} & \multicolumn{1}{c|}{\textbf{\# Tasks}} & \multicolumn{1}{c|}{\textbf{\# Compounds}} & \textbf{Split}& \textbf{Metric} \\
\hline
\multirow{3}[12]{*}{Classification} &  Biophysics   & BBBP  & 1     & 2039  & Scaffold & ROC-AUC \\
\cline{2-6}      & \multirow{3}[10]{*}{Physiology} & SIDER & 27    & 1427  & Random & ROC-AUC \\
\cline{3-6}      &       & ClinTox & 2     & 1478  & Random & ROC-AUC \\
\cline{3-6}      &       & BACE  & 1     & 1513   & Scaffold  & ROC-AUC \\
\cline{3-6}      &       & Tox21 & 12    & 7831  & Random & ROC-AUC \\
\cline{3-6}      &       & ToxCast & 617   & 8575  & Random & ROC-AUC \\
\hline
\multirow{1}[5]{*}{Regression} & \multirow{1}[5]{*}{Physical chemistry} & FreeSolv & 1     & 642   & Random & RMSE \\
\cline{3-6}      &       & ESOL  & 1     & 1128  & Random & RMSE \\
\hline
\hline
\end{tabular}%
\caption{Dataset information.}
\label{tab:dataset_info}
\end{table*}

\begin{table*}[!h]
  \centering
    \begin{tabular}{c|c|c}
    \hline
    hyper-parameter & Description & Value \\
    \hline
    epoch & the number of training epochs. & 20 \\
    $\tau$ & the temperature parameter. & 0.1\\
    batch\_size & the input batch\_size. & 256 \\
    lr & the learning rate.  & 0.0001\\
    GCN\_layers & the number of layers of GCN. & 2 \\
    GCN\_node\_hidden & the node hidden size for GCN. & 64 \\
    KMPNN\_step & the step of message passing for KMPNN. & 6 \\
    KMPNN\_node\_hidden & the node hidden size for KMPNN. & 64 \\
    KMPNN\_edge\_hidden & the edge hidden size for KMPNN. & 64 \\
    node\_out & the node out size for GCN/KMPNN readout. & 64 \\
    edge\_out & the edge out size for GCN/KMPNN readout. & 64 \\
    \hline
    \end{tabular}%
     \caption{The pre-train hyper-parameters.}
  \label{tab:pre-train hyper-parameters}%
\end{table*}

\subsection{Dataset Description}\label{Dataset Description}
Table~\ref{tab:dataset_info} summarizes information of benchmark datasets, including task type, dataset size, split type, and evaluation metrics. We used 6 binary graph classification datasets and 2 binary graph regression datasets. \#Tasks means the number of binary prediction tasks in each dataset. The details of each dataset are listed below \cite{wu2018moleculenet}:
\paragraph{Molecular Classification Datasets.}
\begin{itemize}
    \item \textit{BBBP} \cite{DBLP:journals/jcisd/MartinsTPF12} involves records of whether a compound carries the permeability property of penetrating the blood-brain barrier.
    \item \textit{SIDER} \cite{DBLP:journals/nar/KuhnLJB16} records marketed drugs along with their adverse drug reactions, also known as the Side Effect Resource.
    \item \textit{ClinTox} \cite{gayvert2016data} compares drugs approved through FDA and drugs eliminated due to toxicity during clinical trials.
    \item \textit{BACE} \cite{subramanian2016computational} is collected for recording compounds that could act as the inhibitors of human $\beta$-secretase 1 (BACE-1) in the past few years.
    \item \textit{Tox21} \cite{tox21} is a public database measuring the toxicity of compounds, which has been used in the 2014 Tox21 Data Challenge. 
    \item \textit{ToxCast} \cite{richard2016toxcast} contains multiple toxicity labels over thousands of compounds by running high-throughput screening tests on thousands of chemicals.
\end{itemize}

\paragraph{Molecular Regression Datasets.}
\begin{itemize}
    \item \textit{ESOL} is a small dataset documenting the solubility of compounds \cite{delaney2004esol}.
    \item  \textit{FreeSolv} \cite{mobley2014freesolv} is selected from the Free Solvation Database, which contains the hydration free energy of small molecules in water from both experiments and alchemical free energy calculations.
\end{itemize}

\paragraph{Dataset Splitting.} As shown in Table~\ref{tab:dataset_info}, we apply the scaffold splitting \citep{bemis1996properties} and random splitting as recommended in \cite{wu2018moleculenet} for all tasks on all datasets. Scaffold splitting splits the molecules with distinct two-dimensional structural frameworks into different subsets. It is a more challenging but practical setting since the test molecular can be structurally different from the training set. Here we apply these splitting methods to construct the train/validation/test sets.

\subsection{Baselines}\label{Baselines}
We adopt three types of baselines. The details of each baseline are listed below:
\paragraph{Supervised Learning Methods.}
\begin{itemize}
    \item \textit{GCN} ~\cite{DBLP:conf/iclr/KipfW17} is a convolutional method that focuses on learning the relationship with the nearest neighbor node.
    \item  \textit{Weave}~\cite{DBLP:journals/jcamd/KearnesMBPR16} transformed feature vectors using pair features with distant atoms in addition to atom features focused only on atoms.
    \item \textit{MPNN}~\cite{DBLP:conf/icml/GilmerSRVD17} utilized features from nodes and edges, and summarize it into a framework.
    \item \textit{DMPNN}~\cite{DBLP:journals/jcisd/YangSJCEGGHKMPS19} treated the molecular graph as an edge-oriented directed structure, avoiding the information redundancy during iterations.
    \item \textit{CMPNN}~\cite{DBLP:conf/ijcai/SongZNFLY20} introduced the node-edge interaction module to leverage the node and edge attributes during message passing.
    \item \textit{CoMPT}~\cite{DBLP:journals/corr/abs-2107-08773} invoked a communicative message-passing paradigm based on Transformer. 
\end{itemize}

\paragraph{Pre-trained methods.}
\begin{itemize}
    \item \textit{N-GRAM}~\cite{DBLP:conf/nips/LiuDL19} exploited the idea of N-gram in NLP and conducted vertices embedding by predicting the vertices attributes.
    \item  \textit{Hu et al.}~\cite{DBLP:conf/iclr/HuLGZLPL20}designed two pre-training tasks, i.e., predicting neighborhood context and node attributes, to learn meaningful node representations, then using graph-level multi-task pre-training to refine graph representations.
    \item \textit{GROVER}~\cite{DBLP:conf/nips/RongBXX0HH20} incorporated a Transformer-style architecture and learned node embeddings by predicting contextual properties and motif information.
\end{itemize}

\paragraph{Graph Contrastive Learning Baselines.}
\begin{itemize}
    \item \textit{InfoGraph}~\cite{DBLP:conf/iclr/SunHV020} maximized the mutual information between nodes and graphs.
    \item  \textit{MICRO-Graph}~\cite{DBLP:conf/aaai/Subramonian21} is a motif-based contrastive method.
    \item \textit{GraphCL}~\cite{DBLP:conf/nips/YouCSCWS20} constructed contrastive views of graph data via hand-picking augmentation.
    \item \textit{JOAO}~\cite{DBLP:conf/icml/YouCSW21} automated the augmentation selection in contrastive learning.
    \item \textit{MoCL}~\cite{DBLP:journals/corr/abs-2106-04509} utilized domain knowledge at both local- and global-level to assist representation learning. 
\end{itemize}

\section{Implementation and Pre-training Details}\label{Implementation and Pre-training Details}
We use PyTorch~\cite{NEURIPS2019_9015} and Deep Graph Library~\cite{wang2019dgl} to implement KCL. Table~\ref{tab:pre-train hyper-parameters} demonstrates all the hyper-parameters of the pre-training model. We develop all codes on a Ubuntu Server with 4 GPUs (NVIDIA GeForce 1080Ti).
% We use the  Adam optimizer with learning rate $0.0001$ and batch size of 256. We train the model for 20 epochs. We use GeLU as the activation function. The temperature $\tau$ is set to 0.1. The hidden size for nodes and edges are 64 and 64 respectively. The number of layers of GCN is 2 and the message passing step of KMPNN is 6. We develop all codes on a Ubuntu Server with 4 GPUs (NVIDIA GeForce 1080Ti).

\section{Downstream Details}\label{Downstream Details}
For downstream tasks, we use early stopping on the validation set. We try different hyper-parameter combinations via random search to find the best results. Table~\ref{tab:downstream hyper-parameters} demonstrates  all the hyper-parameters of the fine-tuning model. All fine-tuning tasks are run on a single GPU.

\begin{table}[!h]
\small
  \centering
    \begin{tabular}{c|c|c}
    \hline
    \hline
    hyper-parameter & Description & Range \\
    \hline
    batch\_size & the input batch\_size. & 64,128,256\\
    lr & the learning rate. & 0.0001$\sim$0.1\\
    hidden\_size & the hidden size for predictor. & 32,64,128\\
    patience & the patience for early stopping. & 20\\
    \hline
    \hline
    \end{tabular}%
     \caption{The downstream hyper-parameters.}
  \label{tab:downstream hyper-parameters}%
\end{table}

\section{Additional Experimental Results}
\subsection{Regression Results under Linear Protocol}\label{Regression Results under Linear Protocol}
To be consistent with prior contrastive learning works and make the comparisons fair, we use the linear protocol to evaluate the performance only on classification datasets in main content. Table~\ref{appendix:linear} depicts the additional results of the performance on regression tasks. In order to understand it intuitively, we also listed results under the fine-tune protocol.

\begin{table}[!h]
\small
  \centering
  \begin{tabular}{c|c|c|c}%	\hline
%  \begin{tabular}{c|cccccc}
%    & \multicolumn{6}{c|}{Classification} \\
	\hline
	\hline
    Protocol & Method & ESOL & FreeSolv \\
	\hline
	\multirow{1.5}[2]{*}{Linear} & KCL(GCN) & 0.709 & 1.085\\
	& KCL(KMPNN)  & 0.867  & 1.093\\
	\hline
	\multirow{1.5}[2]{*}{Fine-tune} & KCL(GCN) & \textbf{0.582} & 0.854\\
	& KCL(KMPNN)  & 0.736 & \textbf{0.795}\\	
	\hline
	\hline
	\end{tabular}
	\caption{The performance of KCL under the linear protocol and fine-tune protocol.}
      \label{appendix:linear}
\end{table}

\subsection{Effect of Different Settings}\label{Effect of Different Settings}
We investigate components that influence the performance of the proposed KCL. Table~\ref{tab:ablation-gcn} and Table~\ref{tab:ablation-kmpnn} report specific values of the fine-tuning results in Figure~\ref{ablation}. KCL with knowledge feature initialization and negative sampling scheme shows the best performance among all architectures. Models without all of these two components almost always perform the worst. Excluding any of these two components results in a decrease in performance.

\begin{table}[!h]
\small
\centering
    \begin{tabular}{c|c|c|c|c}
    \hline
    \hline
    \multirow{2}[3]{*}{} & \multicolumn{4}{c}{KCL(GCN)} \\
    \hline
     Dataset & w/oALL & w/oInit & w/oNS & ALL   \\
    \hline
BBBP  & 0.939& 0.941 & 0.953 & \textbf{0.956}   \\
Tox21 &  0.846& 0.853& 0.852 & \textbf{0.856}  \\
ToxCast &  0.750 & 0.751& 0.753 & \textbf{0.757}  \\
SIDER & 0.650 & 0.663 & 0.665 & \textbf{0.666}  \\
CliTox &  0.939& 0.942& 0.940 & \textbf{0.945}  \\
BACE & 0.906 &0.909  & 0.911 & \textbf{0.934} \\
\hline
ESOL & 0.584 &  0.589 & 0.599  & \textbf{0.582}  \\
FreeSolv & \textbf{0.833} & 0.911 &0.896  & 0.854  \\
    \hline
   Ave(Cls) & 0.840 & 0.843 & 0.845 & \textbf{0.852}  \\
   Ave(Reg) & \textbf{0.709} & 0.750 & 0.748 & 0.718\\
   \hline
   \hline
    \end{tabular}%
            \caption{Ablation results on molecular graphs. }
    \label{tab:ablation-gcn}
    \end{table}

 \begin{table}[!h]
    \small
\centering
\begin{tabular}{c|c|c|c|c}
    \hline
    \hline
    \multirow{2}[3]{*}{} & \multicolumn{4}{c}{KCL(KMPNN)}  \\
    \hline
     Dataset & w/oA & w/oInit & w/oNS & ALL   \\
    \hline
BBBP &0.943&0.945&0.956& \textbf{0.961} \\
Tox21 &0.840&0.853&0.856& \textbf{0.859} \\
ToxCast &0.737&0.735&0.739& \textbf{0.740}\\
SIDER &0.650&0.659&0.661& \textbf{0.671}\\
CliTox &0.951&0.956&0.952&\textbf{0.958}\\
BACE &0.919&0.921&0.920& \textbf{0.924} \\
\hline
ESOL & 0.740&0.735&0.739& \textbf{0.732}\\
FreeSolv&0.820&0.798&0.800& \textbf{0.795}\\
    \hline
   Ave(Cls) & 0.840 &0.845 &0.847& \textbf{0.852} \\
   Ave(Reg) & 0.780 &0.765 &0.770&\textbf{0.764}\\
   \hline
   \hline
    \end{tabular}%
    \caption{Ablation results on the augmented molecular graphs. }
    \label{tab:ablation-kmpnn}
    \end{table}

\subsection{Effect of Different Encoders}\label{Effect of Different Encoders}
Since our graph encoder module is pluggable, we explore the impact of different encoders. We replace GCN, KMPNN with GIN and GAT, RGCN and MPNN, respectively. Table~\ref{appendix:ablation} shows the specific values of performance. The results demonstrate that applying different GNN-based encoders on original molecular graphs has no significant impact on performance. Furthermore, KMPNN has a better expressive power on the augmented molecular graphs than the previous heterogeneous GNN and general message passing framework.

\begin{table*}[!h]
\small
  \centering
  \begin{tabular}{c|cccccc|cc}
	\hline
	\hline
    Task & \multicolumn{6}{c|}{Classification} & \multicolumn{2}{c}{Regression} \\
%    & \multicolumn{6}{c|}{(Higher is better)} & \multicolumn{2}{c}{(Lower is better)} \\
	\hline
    Dataset & B\small{BBP} & T\small{ox21} & T\small{oxCast} & S\small{IDER} & C\small{linTox} & B\small{ACE} & E\small{SOL} & F\small{reeSolv} \\
	\hline
	 KCL(GIN) & 0.954   &  0.854   & 0.748  &  0.660 & \textbf{0.945} & 0.932  & 0.580 & 0.856 \\
	 KCL(GAT) &  0.956  &  \textbf{0.857}   & 0.750  & 0.663  & 0.942 & 0.930  & 0.588 & 0.860 \\
	 KCL(GCN) & \textbf{0.956} &  0.856 & \textbf{0.757}  &  \textbf{0.666} & \textbf{0.945}  &  \textbf{0.934} & \textbf{0.582} & \textbf{0.854}   \\
	 \hline
	 KCL(R-GCN) & 0.936  & 0.830 & 0.735  & 0.637  & 0.948 & 0.898  & 0.780 & 1.236 \\
	 KCL(MPNN)  &  0.940  &  0.835   & 0.738  & 0.640  & 0.950 & 0.895  & 0.743 & 1.111 \\
	 KCL(KMPNN)  & \textbf{0.961} & \textbf{0.859} &  \textbf{0.740}  &  \textbf{0.671} & \textbf{0.958}  &  \textbf{0.924} & \textbf{0.732} & \textbf{0.795}   \\
	\hline
	\hline
  \end{tabular}
   \caption{Results comparison with different graph encoders.}
    \label{appendix:ablation}
\end{table*}

\subsection{Effect of Contrastive Learning}\label{Effect of Contrastive Learning}
We investigate the contribution of contrastive learning strategy. Table~\ref{without-cl} depicts the additional results of the comparison of KCL(KMPNN) and KMPNN without contrastive learning under fine-tune and linear protocols.

\begin{table*}[!h]
\small
  \centering
   \begin{tabular}{cccc|ccc}
   \hline
   	\hline
	\hline
    & \multicolumn{3}{c|}{Fine-tune Protocol} & \multicolumn{3}{c}{Linear Protocol}\\
    \hline
	& K\small{CL} & KMPNN & Abs.Imp. & K\small{CL} & KMPNN & Abs.Imp.\\
	\hline
    B\small{BBP} & \textbf{0.961} & 0.915 & +0.046 & \textbf{0.927} & 0.915 & +0.012 \\
    T\small{ox21}  & \textbf{0.859} & 0.804 &  +0.055 & \textbf{0.825} & 0.804 &  +0.021 \\
    T\small{oxCast}  & \textbf{0.740} & 0.725 & +0.015 & 0.709 & \textbf{0.725} & -0.016\\
    S\small{IDER}  & \textbf{0.671} & 0.645 &  +0.026 & \textbf{0.659} & 0.645 &  +0.014\\
    C\small{linTox}  & \textbf{0.958} & 0.892 & +0.066& \textbf{0.898} & 0.892 & +0.006 \\
    B\small{ACE}  & \textbf{0.924} & 0.856 &  +0.068& \textbf{0.860} & 0.856 &  +0.004 \\
    E\small{SOL} & \textbf{0.736} & 0.895 &  +0.159 & \textbf{0.736} & 0.895 &  +0.159 \\
	F\small{reeSolv} & \textbf{0.795} & 2.167 &  +1.372 & \textbf{0.795} & 2.167 &  +1.372\\
	\hline
	Ave\small{(Cls)}  & \textbf{0.852} & 0.806 & +0.046& \textbf{0.813} & 0.806 & +0.007  \\
	Ave\small{(Reg)}  & \textbf{0.765} & 1.531 & +0.766& \textbf{0.766} & 1.531 & + 0.765 \\
	\hline
	\hline
\end{tabular}
  \caption{Results comparison between KCL(KMPNN) and KMPNN without contrastive learning.}
      \label{without-cl}
\end{table*}

Similar to performance under the fine-tune protocol, contrastive learning leads to a performance boost with an average of 0.7\% on classification and 76.5\% on regression over the model without contrastive learning. This reinforces our claim that contrastive learning can incorporate domain knowledge into molecular representations and enhance the prediction performance of downstream tasks.

\subsection{Chemical Interpretability Analysis}\label{Chemical Interpretability Analysis}
\begin{figure}[!ht]
\centering
\includegraphics[width=0.95\columnwidth]{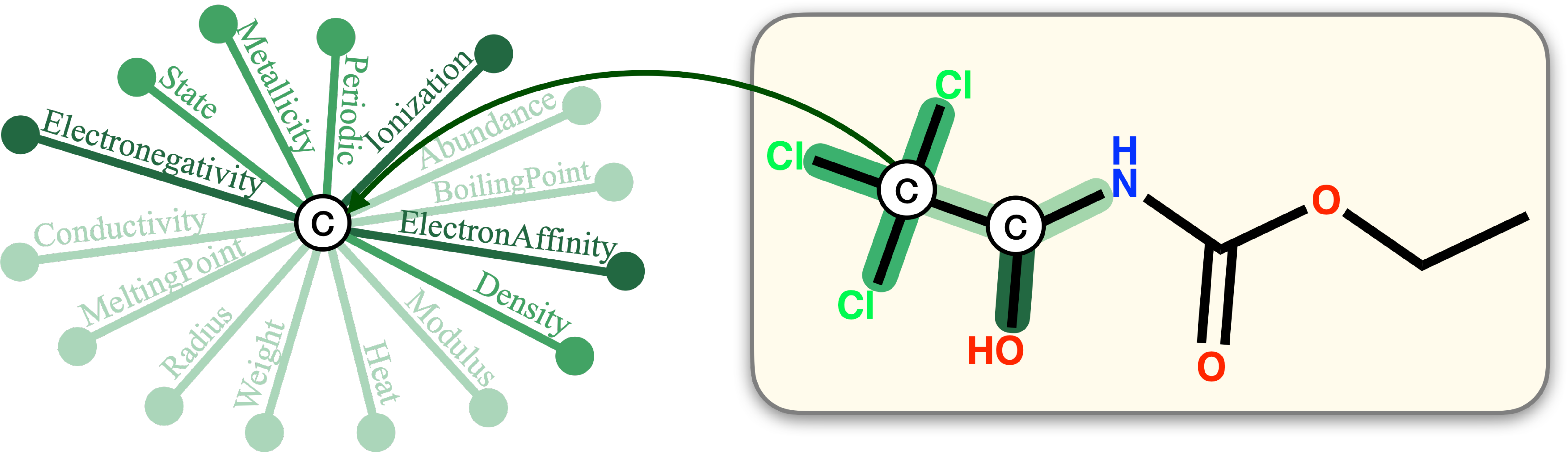} % Reduce the figure size so that it is slightly narrower than the column. Don't use precise values for figure width.This setup will avoid overfull boxes.
\caption{Another example in the BBBP dataset.}
\label{attention1}
\end{figure}

We visualize the attention of each edge in a molecule to explore the interpretability of KMPNN. Figure~\ref{attention1} illustrates another example in the BBBP dataset. Similar to Figure~\ref{attention}, atoms tend to assign more attention to their electron affinity, electronegativity, metalicity, and ionization, which are closely related to atoms' ability to lose electrons. Also, more lively atomic neighbors are easier to be noticed, as shown on the right side of Figure~\ref{attention1}.

\begin{figure}[!h]
\centering
\includegraphics[width=0.95\columnwidth]{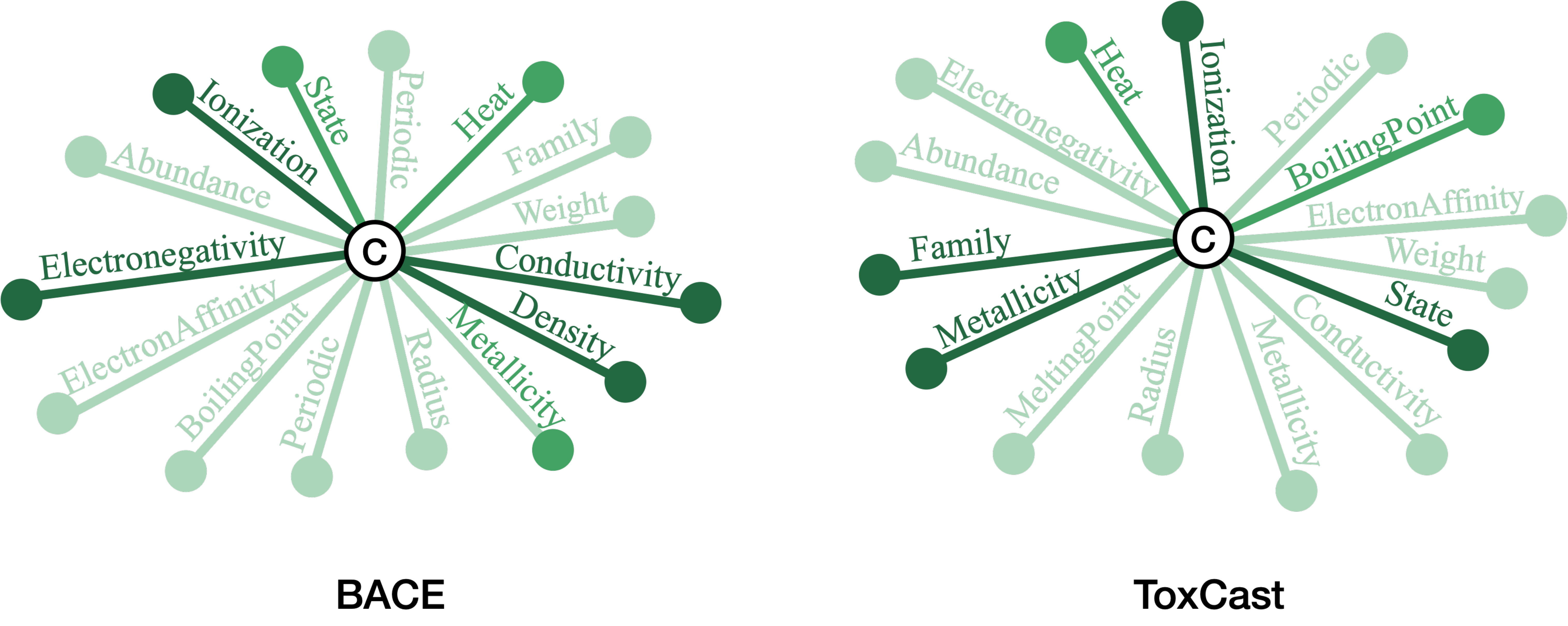} % Reduce the figure size so that it is slightly narrower than the column. Don't use precise values for figure width.This setup will avoid overfull boxes.
\caption{Attention visualization examples of attributes in the BACE and ToxCast datasets.}
\label{attention2}
\end{figure}

Figure~\ref{attention2} shows attention visualization examples on BACE and ToxCast datasets. Most atoms have similar characteristics, here we show two examples of them. Consistent with the conclusion obtained on the BBBP dataset, we observe that fine-grained attributes (e.g., weight, radius) receive less attention than coarse-grained attributes (e.g., electronegativity, conductivity, density, Ionization, metallicity, state, family). This confirms our claim that coarse-grained attributes are more abstract and informative than fine-grained attributes, and therefore contain richer domain knowledge.

\end{document}